\documentclass{article} 
\PassOptionsToPackage{numbers,sort&compress}{natbib}
\usepackage{iclr2027_conference,times}
\setcitestyle{numbers,square,citesep={,}}

\usepackage{amsmath,amsfonts,bm}

\def\eqref#1{equation~\ref{#1}}

\def\1{\bm{1}}

\DeclareMathAlphabet{\mathsfit}{\encodingdefault}{\sfdefault}{m}{sl}
\SetMathAlphabet{\mathsfit}{bold}{\encodingdefault}{\sfdefault}{bx}{n}

\usepackage{hyperref}
\usepackage{url}

\usepackage[utf8]{inputenc} 
\usepackage[T1]{fontenc}    
\usepackage{hyperref}       
\usepackage{url}            
\usepackage{booktabs}       
\usepackage{amsfonts}       
\usepackage{nicefrac}       
\usepackage{microtype}      
\usepackage{xcolor}         
\usepackage{graphicx}
\usepackage{stmaryrd}
\usepackage{bbm}
\usepackage{listings}
\usepackage{subcaption}

\usepackage{mathtools}
\usepackage{algorithm}
\usepackage{algpseudocode}
\usepackage{fancyvrb}

\usepackage{url}
\usepackage{amsmath,amsthm,amssymb,amsfonts}

\title{Learning Mental Models From Unreliable Evidence}

\author{%
  Jeffrey Qin \\
  Computer Science\\
  University of Waterloo\\
  \texttt{jzqin@uwaterloo.ca} \\
  \And
   Zhuangfei Gao \\
   Department of Computer Science \\
   Dalhousie University \\
   \texttt{zhuangfeigao@cmail.carleton.ca} \\
   \And
   Wasu Top Piriyakulkij \\
   Department of Computer Science \\
   Cornell University \\
   \texttt{wp237@cornell.edu} \\
   \And 
   Mia Radovanovic \\
   Department of Psychology \\
   University of Toronto \\
   \And
   Jessica Sommerville \\
   Department of Psychology \\
   University of Toronto \\
   \texttt{jessica.sommerville@mail.utoronto.ca} \\
   \And
   Kevin Ellis \\
   Department of Computer Science \\
   Cornell University \\
   \texttt{kellis@cornell.edu} \\
   \And
   Marta Kryven \\
   Department of Computer Science \\
   Dalhousie University \\
   \texttt{marta.kryven@dal.ca} \\
}

\begin{document}
\raggedbottom

\maketitle

\begin{abstract}
Children expertly construct mental models under uncertainty despite limited world knowledge. Which computational principles make this possible? We study this question in an interactive physical Box Task in which participants infer a latent cause through interaction with a stochastic and partially observed environment. The Box Task was originally designed to study children's puzzle-solving qualitatively. We contribute a computational analysis of behavior that formalizes the cognitive process by which children generate new hypotheses and respond to environmental uncertainty as inference unfolds.
Our modeling framework is based on Bayesian particle-based inference, which maintains a small set of working hypotheses and continually generates new ones.
We give two complementary implementations of hypothesis representations within this framework: (1) as constraints on evidence, and (2) as executable programs evaluated against evidence. Using the constraint-based formulation, we show that the model needs to account both for subjective reliability of evidence and online hypothesis generation to explain key signatures of children’s behavior. 
Using the programmatic formulation, we show that hypotheses may take the form of partial compositional rules to guide information gathering when evidence interpretation is hard.
The programmatic hypotheses are implemented using LLM-based program synthesis, evaluated across three prompt variants and several LLM backends, demonstrating robustness of our approach.
Together, these results extend current models of children's inference to account for subjective reliability of evidence and open-ended hypothesis spaces, and provide a computational analysis of the form generated hypotheses may take in practice.
\end{abstract}


\section{Introduction}

When Newtonian mechanics subsumed Aristotelian physics, it marked a shift from heterogeneous frameworks of motion to a unified theory of inertia and force. Understanding how humans discover such complex models under real-world conditions remains an active challenge in cognitive science~\citep{van2019tasks}, robotics~\citep{collins2024building}, and AI \citep{ying2507assessing}. Developmental studies are particularly informative for uncovering its computational mechanisms, because children construct hypotheses with comparatively limited prior knowledge of the world that can inform model construction~\citep{gopnik2004mechanisms,gopnik2020childhood,gopnik2004theory}.

A central obstacle to the controlled study of real-world model building behavior is the scarcity of experimental frameworks that target naturalistic conditions~\citep{ying2507assessing,allen2024using}. Prior developmental studies use relatively simple tasks in which behavior can be modeled with a constrained hypothesis set ~\citep{cook2011science,denison2013rational,bonawitz2014win}, to demonstrate that children are broadly consistent with approximate Bayesian inference. The Sampling Hypothesis, which proposes that children approximate Bayesian inference by sampling candidate hypotheses from a posterior distribution, is widely established to account for variability in children's responses better than random sampling~\citep{denison2013rational,bonawitz2014win}.
Children also selectively explore informative interventions, and adjust their exploration to both prior causal probabilities and the opportunities to discriminate among likely explanations~\citep{cook2011science}. However, models derived from the Sampling Hypothesis treat evidence as reliable and assume that candidate hypotheses are available in advance, which may not be the case in naturalistic settings.

Moving toward more naturalistic domains, we model behavior in the Box Task \citep{radovanovic2024girls}, an interactive puzzle originally designed to study children's problem-solving qualitatively. The Box Task combines several properties of naturalistic model building including counter-intuitive latent rules that participants are unlikely to consider in advance, partial observability of the environment, 
and an incentive to explicitly model subjective reliability of evidence. 
At a computational level of Marr's analysis \cite{marr1982vision}, this process constitutes a Partially Observable Markov Decision Process (POMDP) given stochastic state transitions and incomplete observations. At the representational level we approximate the intractable POMDP solution \cite{kaelbling1998planning} with particle-based Bayesian inference~\citep{del2006sequential}.

Our modelling framework treats hypotheses as 'particles', allowing two complementary particle representations.
The first, Sets of Constraints (SoC), represents hypotheses as lists of evidence-derived constraints. 
The second, LLM-based Program Synthesis (LLM-PS), represents hypotheses as executable programs generated and revised by an LLM, enabling tractable inference in a flexible hypothesis space \cite{piriyakulkij2024doing,wang2024hypothesis}.
Here SoC functions as a principled model for identifying which cognitive principles drive human behavior, while LLM-PS allows us to examine the form such hypotheses may take when expressed in a Turing-complete high-level language. 
We show that both implementations replicate patterns in children's behavior, demonstrating how the Sampling Hypothesis can be extended to account for open-ended hypothesis spaces and subjective reliability of evidence in practice.

We contribute the following:

\begin{enumerate}

\item \textbf{A formalization of model building behavior} in the Box Task with two complementary implementations of hypotheses as constraints and as programs, extending current models of children's inference.

\item \textbf{A computational analysis of behavior} showing that subjective evidence reliability and online hypothesis generation jointly account for key signatures of children's behavior.

\item \textbf{Performance analysis of LLM-PS}, across prompt variants and  LLM backends, demonstrating the form hypotheses may take and characterizing robustness of our approach. 
\end{enumerate}

\section{Methods}

\subsection{The Box Task}
\label{box_task}

\begin{figure}[t]
    \centering
    \includegraphics[width=\linewidth]{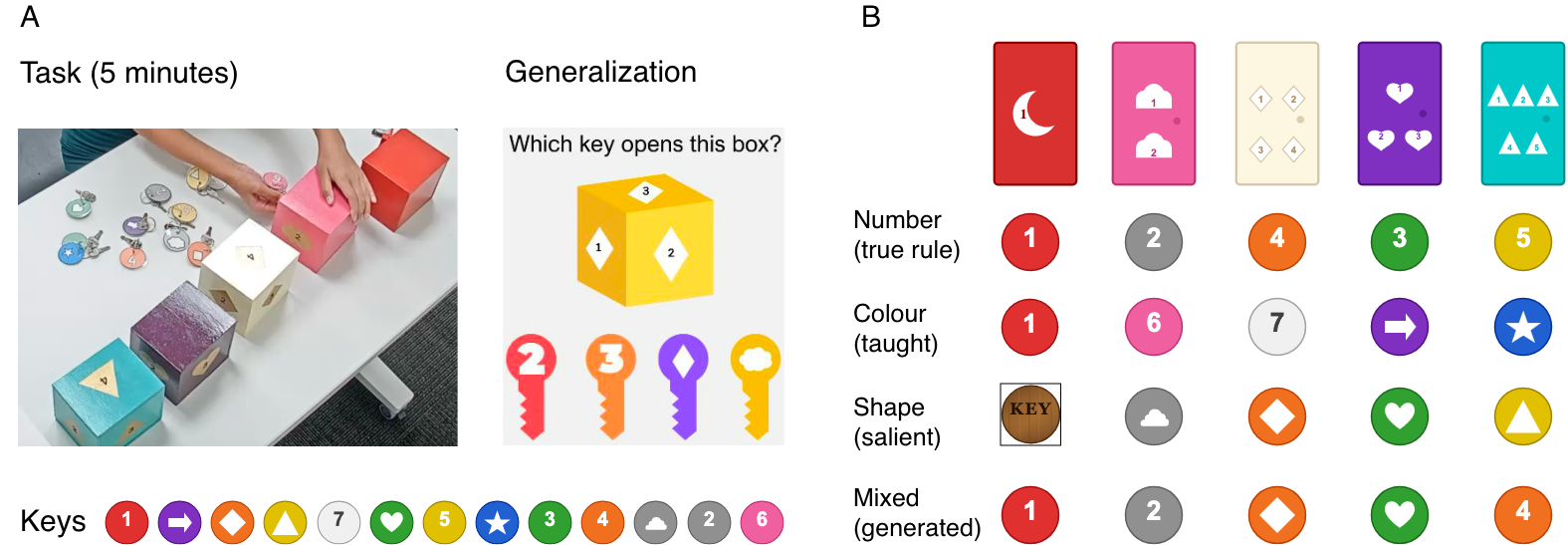}
    \caption{ A. The physical Box Task setup, and the generalization phase presented on a tablet screen. B. Four SoC examples. \textit{Number} shows the ground truth key-box mapping. \textit{Color} represents the taught (misleading) rule. \textit{Shape} rule assumes that any key could open the first box, while other boxes are matched by shape. \textit{Mixed} rule sampled from the online generator selects key-box assignments stochastically. All four rules are consistent with red key opening the red box.}
    \label{fig:task}
\end{figure}

In the Box Task study \citep{radovanovic2024girls} 7-10 year old children ($N=100$) were given 5 minutes to open 5 closed boxes using 13 keys. Boxes are uniquely colored, each marked with multiple instances of a single shape, and presented in a fixed order. Keys have colored fobs displaying either a shape or a number. The true rule, unknown to children, is that each box opens with the key whose number matches the number of shapes on the box (Figure~\ref{fig:task}A). Further, before commencing the task, children are misled that boxes open with color-matching keys. This rule is incorrect, as only the red box can be opened with the red key.
The task is partially observable: not all box faces are visible in the initial line-up, and children are encouraged to pick up the boxes to examine them. 
Following the task, children completed four generalization trials selecting which of four new keys would open an unseen box, where one of the keys corresponds to the true rule and the others are foils. As children may open all boxes without discovering the rule, successful task completion does not entail correct generalization. Although 66\% of children opened all five boxes, only 15\% subsequently generalized the true rule. Full description of the procedure is given in Appendix \ref{si:task}.

\subsection{POMDP formalization}\label{sec:pomdp}

The Box Task constitutes a POMDP, in which the state $s=(n_1,\dots,n_5,o_1,\dots,o_5,h)$ encodes the number of shapes on each box $n_i$, whether each box is open $o_i\in{0,1}$, and the rule (hypothesis) $h$ predicting which key opens each box. The numbers $n_i$ can be observed by picking up the box, $o_i$ is fully observed, and the rule $h$ is inferred during belief updating. At each step, the agent chooses between two action types: \textsc{Observe}$(i)$, which reveals $n_i$, and \textsc{Attempt}$(i,j)$, which attempts to open box $i$ with key $j$.

\textsc{Attempt}$(i,j)$ produces an outcome $y\sim\mathrm{Bernoulli}(\rho^*)$, where $\rho^*$ reflects ground-truth evidence reliability. The action can succeed only if the true rule predicts a match. When $y=1$, the state transitions to $o_i=1$. Since $o_i$ is fully observable, belief is maintained over shape counts and latent rule, $b=p(n_1,\dots,n_5,h\mid\text{history})$.

Exact maintenance of this joint belief over latent state and the open-ended rule space is intractable~\citep{kaelbling1998planning}; we therefore approximate belief inference using particles and use expected information gain as an action-selection heuristic.

\subsection{Probabilistic inference framework}

Our goal is to compute the joint belief distribution $p(h, y, e) = p(y|h, e) p(h) p(e)$, where $h$ denotes latent rule, $y$ denotes outcome, and $e$ denotes action.
We implement this using online Bayesian belief updating over a distribution of hypotheses using Sequential Monte Carlo Samplers (SMC-S;~\citep{Moral_2006}), which is suited to settings where the posterior at each timestep shares a common hypothesis space — as is the case here. 
At each trial $t$, the agent maintains a population of $N$ hypotheses (particles) $\{h^i_t, w^i_t\}_{i=1}^N$, where $h^i_t$ is a hypothesis and $w^i_t$ its weight. 
Here, the number of particles is a free parameter of the model (fewer particles converge faster), which we kept limited to 
a small number in the spirit of resource rationality~\cite{simon1955behavioral}. 
An approximate posterior over hypotheses given $e_{1:t}=(\text{key}, \text{box})$ and outcomes $y_{1:t} \in \{0,1\}$ is:
\[
p(h \mid e_{1:t}, y_{1:t}) \approx \sum_i w^i_t\, \mathbbm{1}[h = h^i_t].
\]

At each step, the agent selects the key-box pair $e^*$ maximizing expected information gain over the current particle distribution 
\citep{lindley1956measure,rainforth2024modern}:
\[
e^* = \arg\max_{e \in \mathcal{E}}\ \sum_i w^i_t\, 
\mathbb{E}_{y \sim p(\cdot \mid e,\, h^i_t)}
\left[D_{\mathrm{KL}}\!\Big(
p(h \mid e_{1:t}, y_{1:t}, e, y) 
\;\Big\|\; 
p(h \mid e_{1:t}, y_{1:t})
\Big)\right].
\]

\paragraph{Belief updating.} Upon observing outcome $y_t$ of action $e_t$, particle weights are updated:
\[
w^i_t \propto w^i_{t-1} \cdot p(y_t \mid e_t, h^i_t).
\]
To model uncertainty over \textit{evidence reliability} as the tendency to attribute failures to external causes (e.g., a stiff lock) rather than immediately falsifying the hypothesis, we introduce a \textit{subjective reliability} parameter $\rho \in [0,1]$.

Then, the likelihood is Bernoulli with success rate $\rho$ when $h$ predicts success and $0$ otherwise:
\[
p(y \mid e, h) = \mathrm{Bernoulli}\!\left(y;\; \rho \cdot \mathbbm{1}[h \text{ predicts } e \text{ succeeds}]\right).
\]

Importantly, while the ground truth key behavior is objectively stochastic, here $\rho$ reflects a subjective estimate of how reliable keys are, supported by empirical evidence that children frequently retry the same failing key--box pair on consecutive trials (Appendix\ref{si:rel}). \textcolor{black}{In our implementation, $\rho$ is fixed throughout each simulation at
$\alpha/(\alpha+\beta)$ for the selected parameter setting (Appendix \ref{si:fitting} );
only the posterior over hypotheses is updated online.}

When the effective sample size  $\mathrm{ESS} = 1/\sum_i (w^i_t)^2$ falls below $N/2$, particles are resampled 
proportionally to their weights and rejuvenated by drawing from the proposal distribution $q(h)$. The rejuvenation mechanism is representation-specific, and described in Sections \ref{sec:soc} and \ref{sec:llmps}. Particles are initialized with weights $w^i_1 \propto p(h^i_1)/q(h^i_1)$. 
In a special case 
where $q(h)$ is configured to track a small set of hypotheses and evidence is treated as reliable, our framework instantiates the Sampling Hypothesis~\cite{denison2013rational} currently accepted as a model of children's rule inference. 

\subsubsection{Hypotheses as Sets of Constraints (SoC)}\label{sec:soc}

Our SoC representation implements each hypothesis $h^i$ as a \textit{Set of Constraints} (SoC): a mapping of boxes to keys predicted to open them. 
Examples of such sets, including the true rule and the taught color rule are shown in Figure~\ref{fig:task} B.
The initial particle population is drawn from $q(h)$, which is implemented as a mixture of two components:

\begin{itemize}
    \item \textbf{Pre-specified rules:} 
    Following the Sampling Hypothesis, we specify a set of candidate rules matching by color, shape, or box order, each assigned a prior probability $p(h)$. 
    \item \textbf{Online generator:} A generative mechanism for proposing novel SoCs not in the pre-specified set, implemented by sampling of key-box assignments conditioned on evidence. Assignments consistent with observed successes are included as a hard constraint, while assignments that have been falsified are down-weighted in proportion to their accumulated failure probability under the current reliability estimate $\rho$.
\end{itemize}

A parameter $p_{\mathrm{gen}}$ controls the proportion of prior mass allocated to the generator versus pre-specified rules. Following the Sampling Hypothesis, the true rule (key number equals number of shapes) is also included in the pre-specified set, with its prior probability $p_t$. In practice, we fixed $p_t = 0.02$ for all model variants reported here, as sweeping $p_t \in [0.01, 0.20]$ consistently resulted in fitted values concentrated near $p_t = 0.02$ across individuals.

\textbf{Model variants. } We construct four SoC variants, including three lesioned versions:

\begin{enumerate}
    \item  \textbf{Lesioned (SoC-Lesioned)}, reliable evidence and no generator ($\rho = 1$, $p_{gen} = 0$). 
    \item  \textbf{Subjective reliability only (SoC-Rel)}, no generator ($p_{gen} = 0$)
    \item  \textbf{Generator only (SoC-Gen)}, no subjective reliability ($\rho = 1$). 
    \item \textbf{Full model (SoC-Full)}, with both parameters $p_{gen}$ and $\rho$ present. 
\end{enumerate}

Special cases of these models instantiate several distinct behaviors.
SoC-Gen with $p_{gen} = 1$ reduces to a random sampler without replacement.
SoC-Full with $p_{gen} = 1$ and $\rho = 0$ reduces to a random sampler with replacement. The SoC-Lesioned formulation effectively instantiates the Sampling Hypothesis. Note that SoC models the \textsc{Attempt} component of the POMDP formulation, but not the \textsc{Observe} policy. The latter is instantiated in LLM-PS-P (see next Section).

\subsubsection{Hypotheses as Symbolic Programs (LLM-PS)}
\label{sec:llmps}

The LLM-PS representation implements hypotheses $h^i$ as executable Python functions \texttt{predict(key, box) -> bool}, evaluated against evidence. The initial particle population is generated by prompting the LLM with a description of the Box Task environment that includes the misleading instruction as a starter hypothesis, and a code template specifying \texttt{Key} and \texttt{Box} data structures. Each particle is an independently sampled program consistent with this context. When a particle $h^i$ is selected for rejuvenation, the LLM is prompted to synthesize a revised version of the corresponding hypothesis consistent with accumulated evidence. This inference machinery determines \textit{which} hypotheses to revise and \textit{when}, while the LLM determines \textit{how} to do so.

\paragraph{Model variants.} We implement three model variants and one REACT baseline:

\begin{enumerate}
\item  \textbf{LLM-PS, full observability, reliable keys}: The LLM receives a complete description of the environment, including full specification of all boxes describing the symbols and numbers on each face. Keys succeed deterministically when attempting to open the correct box. 
\item  \textbf{LLM-PS-S, full observability, unreliable keys}: Evidence reliability is degraded to match empirical success rates experienced by children (see Appendix~\ref{si:rel}). 
\item \textbf{LLM-PS-P, partial observability, unreliable keys}: Box face specification is incomplete in the starter code. The prompt states that the agent may issue \textsc{Observe}$(i)$ actions to reveal them, alongside the regular hypothesis generation. Keys remain unreliable.
\item \textbf{REACT baseline \cite{yao2023react} with full observability and reliable keys}. LLM is prompted sequentially with evidence history and prompted to produce key-box actions without generating explicit hypotheses. This serves as a baseline capable of completing the task, but without producing interpretable hypotheses or generalization to unseen boxes.
\end{enumerate}

The prompts are shown in Appendix \ref{si:prompts}. LLM-PS instantiations vary in LLM backend and number of particles.
Subjective evidence reliability is modeled implicitly by the LLM engine. Parameter $p_gen$ is not modeled directly, as LLM-based models do not separately encode pre-defined vs. stochastically sampled rules. Unlike SoC, LLM-PS rejuvenation is implemented by rejuvenating the worst-fitting hypothesis at every trial, which limits API costs. At task completion, we identify the highest-weight particle in the final particle set. An LLM simulation is classified as \textbf{having generalized} the true rule if this particle implements the correct number-matching rule.
Importantly, we do not treat LLM-PS as a cognitive model of children's behavior. Rather, we use it as a computational tool for open-ended hypothesis generation, to examine the form such hypotheses may take.

\paragraph{Action selection.}
In LLM-PS-P, the model can issue \textsc{Observe}$(i)$ actions to update the environment context passed to subsequent LLM calls. Particles generated prior to observation retain their original representation and may be discarded through subsequent rejuvenation; particles generated after observation can exploit the revealed information about box layout. This implements a factorized approximation to the joint belief $p(n_{1:5}, h \mid \text{history})$, treating state uncertainty and rule uncertainty independently in a tractable approximation of the formalism described in Section~\ref{sec:pomdp}. 

\begin{figure}[t]
    \centering
    \includegraphics[width=1\linewidth]{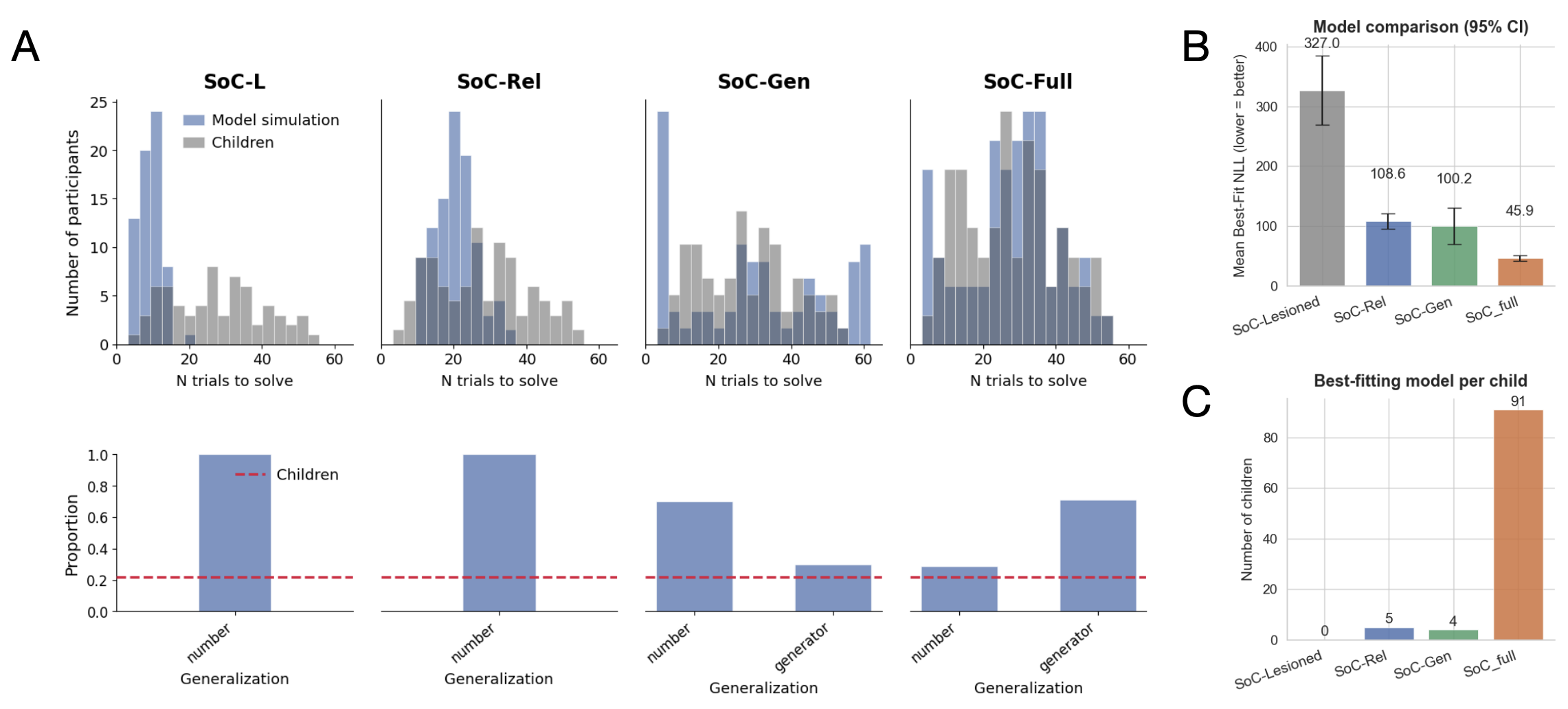}
    \caption{A, Top row. Histograms showing the number of trials required to open all boxes for children and models. Each column represents a different model variant. The same children's histogram is repeated in each column as a reference. Bottom row: Distribution of most likely hypotheses for each model after opening all boxes. The correct rule, "number", was generalized to by 15\% of children. B. Negative Log-likelihoods of each model (shorter = better fit). Error bars indicate 95\% CI computed via standard error of the mean across children. C. Number of children best fitted by each model. }
    \label{fig2}
\end{figure}

\section{Experiments}

\subsection{Children's Modeling Results (SoC) }

We first evaluate the SoC model's ability to capture (1) an individual's task-completion trajectory, defined as cumulative boxes opened over trials, 
and (2) the final inferred hypothesis. We fit each model variant to individuals, with parameter selection maximizing the composite log-likelihood of that child's trajectory under each model (Appendix~\ref{si:fitting}), and report the best fitting model per child (Figure \ref{fig2}C) along with the population mean LL for each model (Figure \ref{fig2}B). The SoC-Full model provides a significantly better fit than all lesioned variants ($p < .001$): SoC-Rel ($t=13.39, d=1.34$), SoC-Lesioned ($t=10.14,d=1.01$), and SoC-Gen ($t=3.90, d=0.39$). AIC-based model comparison confirms this ranking after penalising for model complexity (Appendix~\ref{si:fitting}, Table~\ref{tab:aic}). 

Figure~\ref{fig2}A shows histograms of trials required to open all five boxes for each model variant, given the fitted parameters, alongside children's distribution. Only the SoC-Full model closely reproduces the shape and spread of children's aggregate distribution. The bottom row of Figure~\ref{fig2} shows the distribution of final inferred rules. The lesioned models without the online generator (SoC-Lesioned and SoC-Rel) can only generalize to pre-defined rules, forcing them to consistently pick the true rule. Models with the online generator (SoC-Gen and SoC-Full) can complete the task without discovering the true rule, converging on a set of independent constraints for each box, labeled 'generator' in the figure. The SoC-Full model converges on the correct rule in approximately 20\% of simulations, similar to children's generalization. 

\textcolor{black}{To evaluate whether the fitted models also predict children's individual exploratory choices, we conducted an additional action-level predictive analysis (Appendix~\ref{si:action_validation}).
Using the previously fitted parameters, we conditioned each model on the child's observed action and outcome history and evaluated its predictive distribution over subsequent key--box choices.
The action-level analysis further supports the contribution of online hypothesis generation: SoC-Full significantly outperformed SoC-Rel ($t(99)=9.38$, $p<.001$) and SoC-Lesioned
($t(99)=10.25$, $p<.001$), but its improvement over SoC-Gen
was not statistically significant ($t(99)=1.00$, $p=.319$).
These results show that the advantage of models incorporating
online hypothesis generation extends beyond predicting cumulative
task progress to predicting individual exploratory choices.
The additional improvement from subjective reliability was
not significantly established in this analysis.}

\begin{figure}[t]
    \centering
    \includegraphics[width=0.9\linewidth]{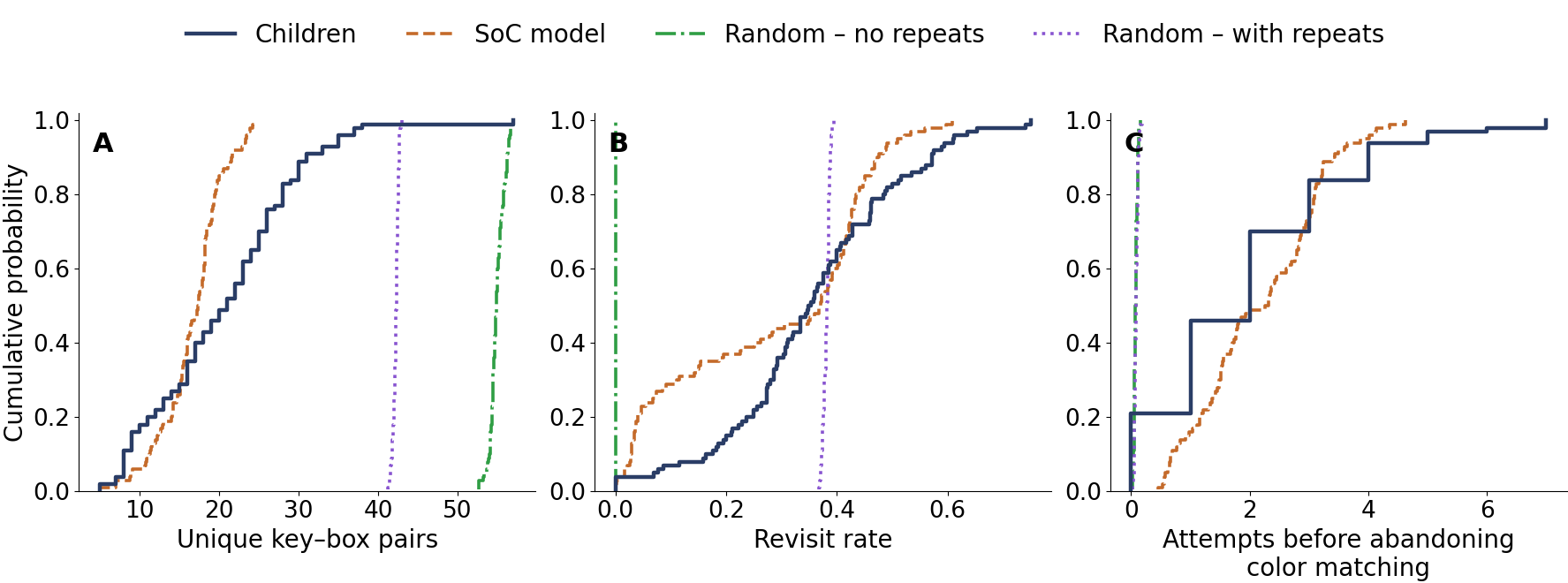}
    \caption{
    \textbf{Comparison of observed behavior with the SoC model and random-search baselines.}
    Distributions of (A) Unique key-box pairs attempted (B) key-box revisit rate and (C) attempts before abandoning color matching. Behavioral data (solid line) is compared to our model (shown in red), random sampling without replacement (green), and random sampling with replacement (blue).}
    \label{fig:behavioral_baselines}
\end{figure}

Lastly, children's behavior cannot be explained by a fully stochastic process, such as random sampling without or without replacement. Notably, the special case of SoC-Gen with maximal $p_{gen}$ value that reduces to random sampling without replacement shows a poor fit to behavior (see Appendix~\ref{si:fitting} and Figure \ref{fig:LL_surfaces}). The special cases of SoC-Full with maximal value of $p_{gen}$ and various values of $\rho$, that enable key resampling, is likewise a poor fit. We additionally compare our model predictions with random-sampling baselines either with or without replacement (Fig.~\ref{fig:behavioral_baselines}) on three metrics: how many unique key-box pairs were attempted, key-box revisit rate, and attempts before abandoning color matching, demonstrating that only our model captures the empirical metrics. Additional evidence against a random-sampling account provided by children's initial bias toward testing the instructed color rule is shown in (Appendix\ref{si:behavior} Figure~\ref{fig:attempts-before-abandoning-color}).  

Finally, the majority of children express a significant rule preference during generalization, by above-chance picking key-box combinations consistent with a specific rule; and only 26 did not show such a preference (Appendix \ref{si:behavior}, Figure \ref{fig:generalization-histogram}). 
Taken together these results establish that subjective reliability ($\rho$) and online hypothesis generation ($p_{\mathrm{gen}}$) are jointly required to account for children's behavioral signatures.

\subsection{Simulation Results (LLM-PS)}

\textbf{LLM-PS performance across backends.} 
We evaluated the three LLM-based model variants with three LLM backends, with different architectures and code generation capabilities (\textsc{gpt-5.2, low reasoning}, \textsc{gpt-5.1}, \textsc{DeepSeek V3.2}). 
\textsc{DeepSeek V3.2} and \textsc{gpt-5.1} were instantiated with $n=10$ particles, as a sufficient number at which the task could be consistently completed in under 70 attempts -- the maximal number in behavioral data. We also found that a more capable model, \textsc{gpt-5.2}, was able to complete the task with a single particle. In the single-particle limit, LLM-PS reduces to sequential hypothesis revision: the agent returns a single hypothesis, used by the inference engine to greedily select actions; the rejuvenation step prompts to update this hypothesis given history. Empirically, this simplification is consistent with the view that capable inference systems can operate effectively under a single-hypothesis regime \cite{bramley2017formalizing}. At the computational level, one-particle LLM-based models with a \textsc{gpt-5.2} back-end can be seen as an instantiation of the Win-Stay, Loose-Sample framework \cite{bonawitz2014win} with interpretable hypotheses.

\begin{figure}[t]
    \centering
    \includegraphics[width=\linewidth]{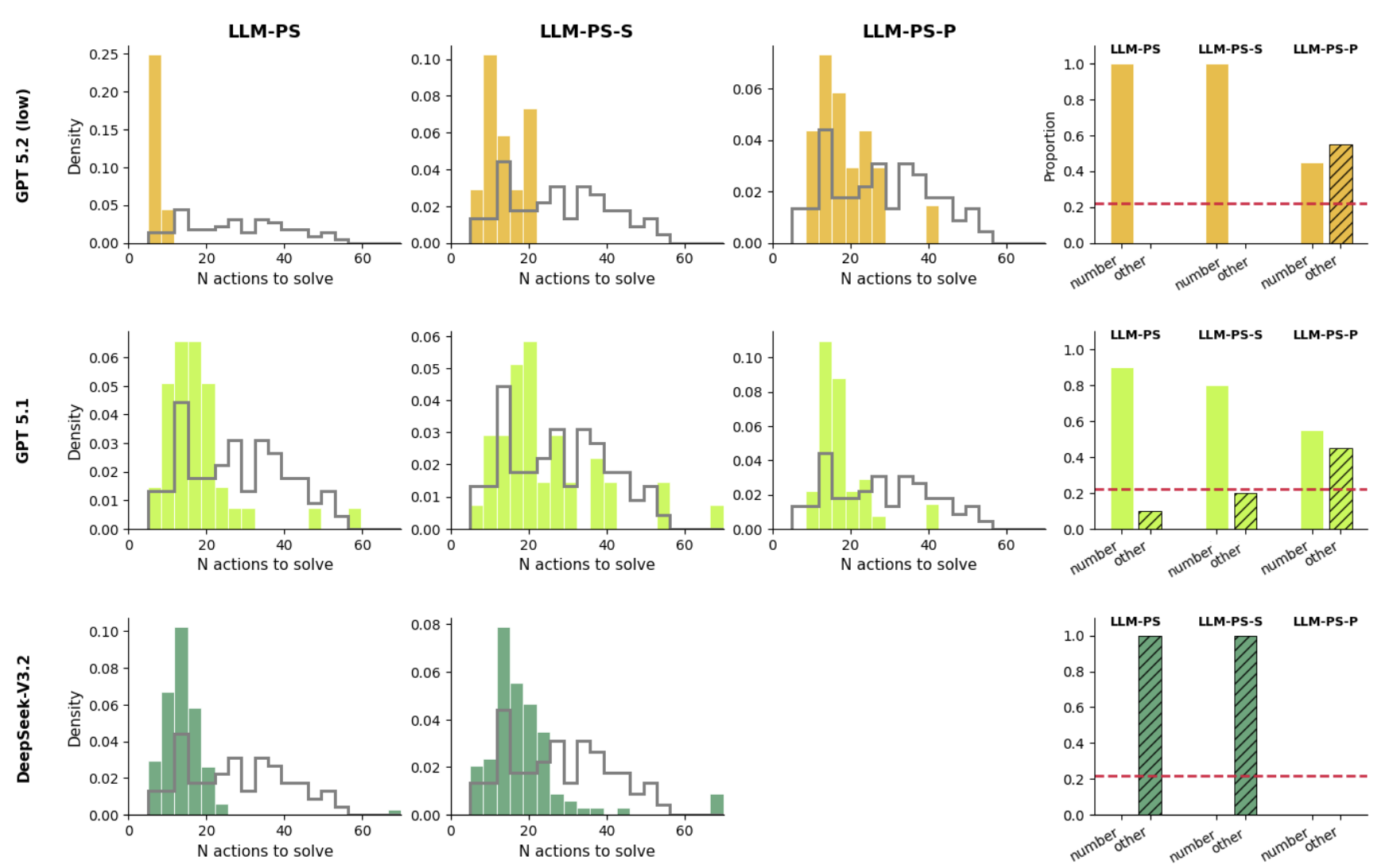}
    \caption{Left: Densities of number of attempts required to open all five boxes under each LLM-PS variant, across LLM backends (rows). Columns correspond to model variants. 
    Right: Generalization of the true rule by each model variant. The red dotted line shows children's generalization from \cite{radovanovic2024girls}.}
    \label{fig:llmps_results}
\end{figure}

Figure~\ref{fig:llmps_results} shows the number of actions required to complete the task across LLM-based model variants and LLM backends. The ReAct baselines (not plotted) consistently completed the task in under 15 attempts across backends, confirming that under reliable and fully observable 
conditions task completion does not require explicit hypothesis maintenance. Here, ReAct simply provides a task completion baseline, without explicit hypothesis representations.
Given that \textsc{DeepSeek V3.2} failed to discover the true rule under full observability (LLM-PS and LLM-PS-S), it was not further evaluated under LLM-PS-P. 

\textbf{LLM-PS performance across prompt variants.} 
Example trajectories for two model variants using \textsc{gpt-5.2} are shown in Appendix \ref{si:prompts} Figure \ref{fig:hypothesis_trajectories}. Both start with a color-matching hypothesis, consistent with the misleading instruction, but converge on the correct rule. 
Similar to children, LLM-based models dissociate task completion and generalization under increasing uncertainty. 
When the model fails to discover the correct rule despite completing the task, the failure modes were similar across LLMs. In such cases, task success was achieved by accumulating coincidental successful box openings driven by hypotheses that express disparate causes, rather than unified rules (Appendix \ref{si:prompts}), a property also captured by the online generator in the SoC. 
Notably, the weaker model, \textsc{DeepSeek V3.2}, consistently failed to discover the true rule even in fully observable conditions.
These results demonstrate how LLM-based proposals in SMC-S can produce interpretable hypotheses based on evidence history, demonstrating the form hypotheses might take during online generation.

\textbf{Key differences between children and LLM-based proposal behaviour.}
Across prompt variants, LLM-based proposals strictly complied with instructions, consistently generating the color-matching hypothesis on the first trial. In contrast, 21 out of 100 children tested a key-box combination inconsistent with color matching, suggesting that children may have competing inductive biases or assume a broader hypothesis space. This property of children's behavior is captured by SoC, but not by LLM-based proposals. Further, in contrast to SoC, LLM-PS-P supports both \textsc{Observe} and \textsc{Attempt} actions. Our instantiation of LLM-PS-P exhibits a more systematic observation policy than children (Appendix Figure~\ref{fig:si_observe_actions}), who vary substantially in how and when they observe boxes: 30 out of 100 did not observe any boxes; the majority (79 out of 100) examined only a subset; and 22 out of 100 began the experiment by observing boxes rather than attempting keys. In contrast, LLM-PS-P consistently begins by generating \textsc{Attempt} actions and switches to systematic \textsc{Observe} actions after a few failed attempts (Appendix~\ref{si:example_log}). The differences between LLM-based proposals and children's behavior may be due to observations differing in informativeness, cost, or both. 

\textcolor{black}{ We additionally compare LLM-PS exploration with children's behavior using the same three descriptive measures employed
in Figure~3: unique key--box pairs attempted, revisit rate,
and persistence with the instructed color-matching strategy
(Appendix~\ref{si:llm_behavior}). These comparisons further
characterize LLM-based exploration across reliability and
observability conditions, without implying fitted
predictions of children's behavior. }

\begin{figure}[t]
\centering
\includegraphics[width=\linewidth]{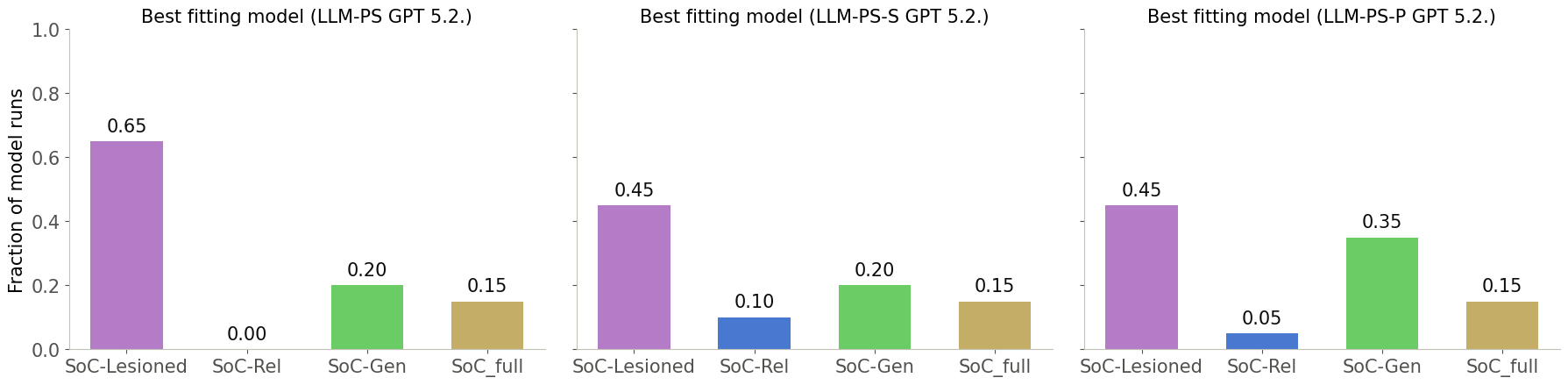}

\vspace{0.8em}

\includegraphics[width=\linewidth]{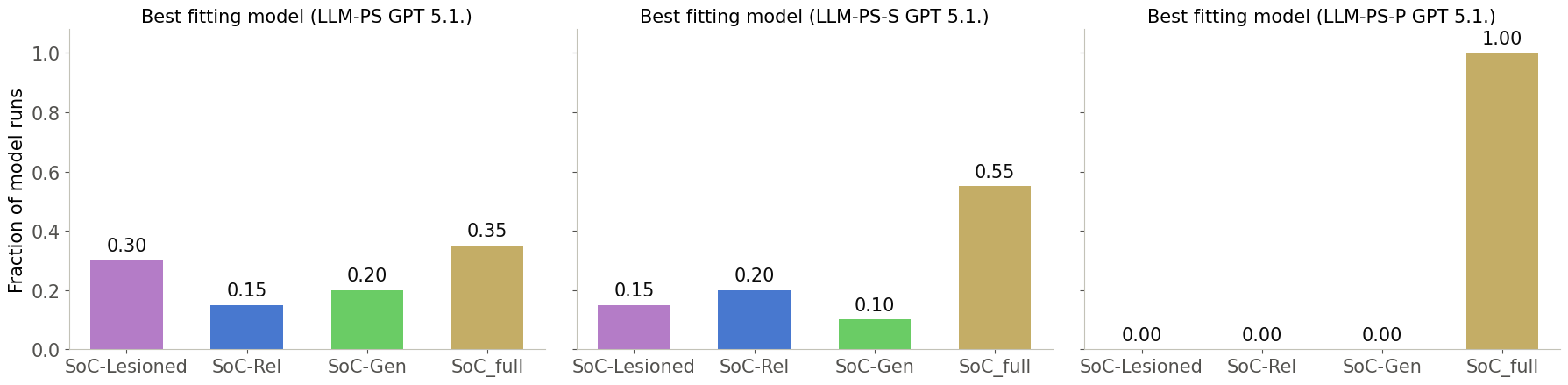}
\caption{Best-fitting SoC variants for \textsc{Attempt} actions generated by LLM-PS with GPT~5.2 (top) and GPT~5.1 (bottom). Bars show the fraction of runs assigned to each SoC variant.}
\label{fig:best-fitting-model-all}

\end{figure}

\textbf{Behavioral correspondence between LLM-based and SoC proposals.}
Unlike SoC proposals, whose behavior is explicitly parameterized by $\rho$ and $p_{\mathrm{gen}}$, LLM-based proposals do not expose corresponding parameters. Their differences in generated hypotheses arise implicitly from the LLM and its prompt. To characterize this implicit behavior, we fit the SoC models to \textsc{Attempt} actions generated by each LLM-PS variant (Figure \ref{fig:best-fitting-model-all}). This provides a mapping from the LLM-based proposal mechanisms onto the interpretable SoC parameterization. Across LLM backends, under fully observable, reliable conditions, LLM-PS maps most closely onto SoC-Lesioned, whereas introducing unreliable evidence (LLM-PS-S) and partial observability (LLM-PS-P) progressively shifts behavior toward the SoC-Full. 

\section{Related Work}

\paragraph{Bayesian models of learning.}
Bayesian inference has been widely applied to model child \citep{bonawitz2014probabilistic,gopnik2020childhood} and adult causal learning \citep{griffiths2012bayesian,lake2015human,rule2020child,kemp2009structured}.
Most directly related, the Sampling Hypothesis proposes that children sample causal hypotheses from a fully predefined proposal distribution weighted by posterior probability \citep{denison2013rational,bonawitz2014win,vul2014one}. Our SoC model 
extends this principle to complex domains where evidence is unreliable and the proposal cannot be entirely predefined. In other related work, \citet{bramley2023active} find that children generate more diverse hypotheses but converge on correct rules less reliably than adults, a pattern consistent with our finding that children's hypothesis space admits solutions that succeed at task completion without capturing the true rule.

\paragraph{Program synthesis for inductive reasoning.}
Much work across AI and cognitive science uses program synthesis to model inductive reasoning.
Classically such models have had relatively narrow success because they use custom domain-specific programming languages~\cite{10.1145/1925844.1926423,balog2016deepcoder,goodman2008rational,lake2015human,piantadosi2011learning,DBLP:journals/corr/abs-1911-01547,10.1145/2736282}, although many of those successes have been practically and scientifically valuable.
Recent work with general-purpose languages such as Python using LLMs has increased the scope of applications~\citep{piriyakulkij2025poe,dainese2024codeworldmodel, wang2024hypothesis,qiu2024phenomenal,10.5555/3737916.3739338}.
Most closely 
related, \citet{piriyakulkij2024doing} also combines SMC-S and Bayesian Experiment Design with LLM-based proposal distributions. Our work differs in core ways, but most importantly in its comparison to children and SoC models, and in its use of a naturalistic task domain in which task completion and model learning can be dissociated. Notably, our finding that \textsc{gpt-5.2} completes the task with a single particle is consistent with the view that capable inference systems can operate effectively under a single-hypothesis regime \citep{bramley2017formalizing,bonawitz2014win}.
Our view is that programs induced from naturalistic data should not assume full observability, and that decision-making must consider uncertainty about both the program hypothesis and the underlying state.
Indeed, absent this modeling choice, we cannot explain the human data we consider.


\section{Discussion}

We introduced a computational framework modeling the behavior of children interacting with the Box Task, which expresses several key properties of model building in a naturalistic setting: evidence uncertainty, partial observability, and non-intuitive causal rules.
Our framework formalizes behavior in this task as the process of hypothesis generation, interpreting unreliable evidence, and sequential belief revision. A key component of our framework is the mechanism that specifies how new hypotheses may be generated given evidence history. We implement this component in two complementary ways: the SoC class of models, which extends the Sampling Hypothesis \citep{denison2013rational} to uncertain domains with open-ended hypothesis spaces, and a program synthesis LLM-PS class of models, which enables tractable inference over expressive, interpretable, and executable hypotheses.

Using SoC lesion analysis we show that evidence uncertainty and online hypothesis generation are jointly needed to account for children's behavior. Additional LLM-based modeling demonstrates the form that these hypotheses might take in an uncertain environment: compositional or partial rules, that specify the underlying causal model only partly. Our modeling framework shows the dissociation between task completion and correct generalization similar to children, as both LLM-based and SoC model families can generate hypotheses that integrate disparate causes rather than propose new unified rules. We show that such hypotheses can be instrumental both to completing the task without correct generalization, and to accumulating evidence, which can then inform generation of improved unified rules.
The extent to which human-generated hypotheses may follow a similar pattern from cause aggregation to unification remains to be tested in future work. 


We find that LLM-based hypothesis generation handles unreliable evidence gracefully, but appears to work with a fairly restrictive prior over hypotheses. Generated hypotheses primarily match keys and boxes by simple salient properties: shape identity, color, and ultimately number of shapes. In our simulation experiments, we did not encounter LLMs coming up with plausible but counter-intuitive rules, for instance \textit{'The number on the key needs to be odd and greater than the position of the box in the line-up'.} or \textit{'The rule changes after completing 10 attempts.'}. This focus on simple rules explains the efficiency of LLM-based model performance. Possibly, training on natural language descriptions of experiment-like tasks induces an LLMs bias toward small hypothesis spaces, and training on noisy data (i.e. unreliable witnesses) could inform discounting conflicting evidence.   

Notably, as the Box Task originates from a published study~\citep{radovanovic2024girls}, LLM 
training corpora may have included the task design or its solutions. 
While contamination cannot be fully excluded, the observed trajectories consistently generate the simpler salient hypotheses before discovering the true rule and demonstrate evidence-sensitive revision conditional on the proposals produced by the LLM. 
Moreover, weaker models fail to generate the true rule entirely. 

\textcolor{black}{
While LLM-PS generated both unified and partial, compositional
rules, we did not observe unrelated key--box assignments of the
kind supported by the SoC generator, which we found to be important for explaining children's behavior. In the fully observable LLM-PS conditions, the prompts explicitly expose numerical attributes of keys and boxes, making the true number-matching rule salient and likely contributing to the ease
with which the models discover it. In LLM-PS-P, box attributes
are initially hidden, but the agent systematically observes
boxes after unsuccessful attempts, revealing information that
children acquire more selectively and potentially incompletely.
This distinction is particularly relevant in light of previous
work by Piriyakulkij et al.~\cite{piriyakulkij2024doing},
which employs LLM-guided probabilistic inference to model human
rule learning in a Zendo-style task. Given the differences
between LLM-based exploration and children's behavior, as well
as the influence of task specification on generated hypotheses,
we caution against interpreting LLM-PS as a cognitive model of
children's inference. Rather, our aim is to examine how
LLM-based program synthesis can support the generation and
revision of interpretable hypotheses within an open-ended
hypothesis space. }

\textcolor{black}{ Our results suggest that the number of particles required for effective inference may depend on the capabilities of the LLM proposal mechanism. While GPT-5.2 successfully completed the task with a single particle, pilot experiments with
GPT-5.1 and DeepSeek-V3.2 were less reliable, requiring approximately 10 particles for consistent task completion. While we did not systematically vary particle count, our results suggest that combining potentially cheaper LLMs with
multi-particle inference may be cost-effective. }

\paragraph{Limitations and Future work}
The assumption that \textsc{Observe}$(i)$ actions in LLM-PS-P are fully informative could be relaxed by introducing noisy observations in future work. Future work could also consider a mechanistic investigation of whether LLM-based agents may maintain explicit hypothesis representations using open-weight models. Follow-up experiments with human adults could subject human adults and LLM-based models to matched environments (fully observed and reliable; fully observed and stochastic; partially observed and stochastic) while eliciting explicit descriptions of hypotheses from human adults. Doing so would inform comparative analysis of hypotheses generated by LLM-based models and humans, offering insight into how the underlying hypotheses are similar or different in quality and diversity between human and artificial agents. 






\section*{AI Use Statement}
LLMs were used as part of the methodology described in this paper, specifically as proposal mechanisms for structural decompositions and reconstruction programs in the Generative Map Module; these uses are described in detail in the Methods and Appendix. 

Generative AI tools were also used for routine coding assistance with generating plots, proofreading the completed manuscript, and literature discovery. All AI-assisted content and suggested references were reviewed and verified by the authors, who take responsibility for the final content of the paper.

\medskip
\bibliographystyle{iclr2027_conference}

{
\small
\bibliography{refs}
}


\appendix

\section{The Box Task Setup}\label{si:task}

The Box Task environment is introduced in \citep{radovanovic2024girls}. Below we summarize relevant  details. 

Children completed the experiment in person, in four phases:

\begin{enumerate}
\item \textit{Practice}: Participants tried plain keys on a plain practice box to familiarize themselves with the key action, including both correct and incorrect keys. 
\item \textit{Instruction}: Children viewed an instructional video in which a teacher demonstrated an incorrect color-matching rule, using a red-fobbed key to open the red box. This demonstration was confounded: for the red box only, the correct key happened to match both color and number. For all other boxes, color-matched keys were incorrect. The teacher communicated the rule as definitive and provided no indication that it would fail.
\item \textit{Test}: The five boxes were arranged in a fixed order in front of the participant. All 13 keys were placed in a single pile. Children were instructed to open all five boxes within a 5-minute time limit, working independently. The experimenter remained present but provided no feedback. Boxes could be picked up and examined. The test ended upon opening all five boxes or at time expiry.
\item \textit{Generalization}: Four forced-choice trials using novel box images were presented on a tablet screen. For each box, four keys were presented: a color-matched key, a shape-matched key, a number-matched key (correct), and a number foil. Children selected which key they believed would open the box.
\end{enumerate}

\paragraph{Box Task Environment.}

The boxes were uniquely colored physical objects, 11 cm $\times$ 11.9 cm $\times$ 10.6 cm, featuring a different number of shapes on their faces. Each shape was labeled with a number to indicate quantity (e.g., the pink box had two clouds, labeled "1" and "2"). Each box had at least one shape.
Keys were metal physical objects. Each key had a colored tag displaying either a number or a shape. 

\paragraph{Partial Observability.}
The task is partially observable: subjects initially observe each box's color and 
shape identity, but not the total number of shape instances. The 
exact shape count is revealed only if the subject picks up and examines the box.

At each time $t$, subjects choose between:
\begin{itemize}
    \item \textbf{Observe:} $\text{observe}(i)$ reveals the exact number of shapes on box $i$
    \item \textbf{Attempt:} $\text{attempt}(i,j)$ attempts to open box $i$ with key $j$, where $i \in \{1,\dots,5\}$ and $j \in \{1,\dots,13\}$
\end{itemize}


\begin{table}[ht]
\centering
\begin{tabular}{lllll}
\hline
Box & Order & color & Shape & Shape count (Number) \\
\hline
B1 & 1 & red & moon & ? \\
B2 & 2 & pink & cloud & ? \\
B3 & 3 & white & diamond & ? \\
B4 & 4 & purple & heart & ? \\
B5 & 5 & blue & triangle & ? \\
\hline
\end{tabular}
\caption{Initial box configuration as observed by subjects. Color, shape identity, and order are fully observable. Shape count requires picking up the box.}
\end{table}

\begin{table}[ht]
\centering
\begin{tabular}{llll}
\hline
Key & color & Shape & Number \\
\hline
red1 & red & & 1 \\
pink6 & pink & & 6 \\
grey2 & grey & & 2 \\
greycloud & grey & cloud & \\
orange4 & orange & & 4 \\
green3 & green & & 3 \\
bluestar & blue & star & \\
yellow5 & yellow & & 5 \\
greenheart & green & heart & \\
white7 & white & & 7 \\
triangleyellow & yellow & triangle & \\
diamondorange & orange & diamond & \\
purplearrow & purple & arrow & \\
\hline
\end{tabular}
\caption{Set of available keys. Each key is associated with a color and either a number or a shape (but not both). Numbers range beyond the possible box counts, introducing distractors; shape-based keys provide alternative matching cues.}
\end{table}

\begin{table}[ht]
\centering
\begin{tabular}{lllll}
\hline
Box & Order & color & Shape & Number \\
\hline
B1 & 1 & red & moon & 1 \\
B2 & 2 & pink & cloud & 2 \\
B3 & 3 & white & diamond & 4 \\
B4 & 4 & purple & heart & 3 \\
B5 & 5 & blue & triangle & 5 \\
\hline
\end{tabular}
\caption{True (unobserved) configuration of the boxes. The true number of shapes on each box becomes observable with inspection.}
\end{table}

\clearpage

\section{POMDP Formalization}\label{si:pomdp}

The Box Task constitutes a Partially Observable Markov Decision Process, in which the state $s = (n_1,\dots,n_5, o_1,\dots,o_5, h)$ encodes the number of shapes on each box $n_i$, whether each box is open $o_i \in \{0,1\}$, and the rule (hypothesis) $h$ predicting which key opens each box.  The numbers $n_i$ can be observed by picking up the box, $o_i$ is fully observed, and the rule $h$ can be inferred during belief updating. The reward $r = 1$ when $\forall i~ o_i = 1$, and $0$ otherwise. The goal state is reached when all boxes are open, $\sum_i o_i = 5$. At each step, the agent chooses between two action types: $\textsc{Observe}(i)$: deterministically reveals information about $n_i$ and $\textsc{Attempt}(i,j)$: attempts to open box $i$ with key $j$.

\textsc{Attempt}(i,j) produces an outcome $y \sim \text{Bernoulli}(\rho*)$, where $\rho*$ reflects ground truth evidence reliability. The action can succeed only if the oracle rule predicts a match. When $y=1$, the state transitions to $o_i = 1$. Since $o_i$ is fully observable, the belief is a joint distribution over shape counts and latent rule: $b = p(n_1,\dots,n_5,\, h \mid \text{history})$. Updating $h$ after taking an action requires computing likelihood over possible rules.  

We use the POMDP to formalize the environment and latent state. As exact maintenance of the joint belief $p(n_{1:5}, h \mid \text{history})$ is intractable~\citep{kaelbling1998planning}, we approximate belief inference using particles and use expected information gain as an action-selection heuristic.

\section{LLM Prompts and Output Examples}\label{si:prompts}

\begin{figure}[th]
\centering
\begin{minipage}[t]{0.48\textwidth}
\centering
\textbf{LLM-PS-S (Unreliable evidence)}
\vspace{0.3em}

\begin{lstlisting}[language=Python, basicstyle=\ttfamily\scriptsize,
  frame=single, title={\small\itshape Trial 1: Color rule (misleading prior)}]
def predict(key, box) -> bool:
    # Match key color to box color
    return key.color == box.color
\end{lstlisting}
\vspace{0.3em}

\begin{lstlisting}[language=Python, basicstyle=\ttfamily\scriptsize,
  frame=single, title={\small\itshape Trial 6: Partial revision}]
def predict(key, box) -> bool:
    # color for first box, then shape
    if box.position == 1:
        return key.color == box.color
    return (key.shape is not None and
            key.shape == box.shape)
\end{lstlisting}
\vspace{0.3em}

\begin{lstlisting}[language=Python, basicstyle=\ttfamily\scriptsize,
  frame=single, title={\small\itshape Trial 14: True rule}]
def predict(key, box) -> bool:
    # Key number must match the number
    # of shapes on the box.
    return (key.number is not None and
            key.number in box.number)
\end{lstlisting}

\end{minipage}
\hfill
\begin{minipage}[t]{0.48\textwidth}
\centering
\textbf{LLM-PS-P (Partial Obs., Unreliable)}
\vspace{0.3em}

\begin{lstlisting}[language=Python, basicstyle=\ttfamily\scriptsize,
  frame=single, title={\small\itshape Trial 1: Color rule (misleading prior)}]
def predict(key, box) -> bool:
    # Match key color to box color
    return key.color == box.color
\end{lstlisting}
\vspace{0.3em}

\begin{lstlisting}[language=Python, basicstyle=\ttfamily\scriptsize,
  frame=single, title={\small\itshape Trial 9: Uncertain number belief}]
def predict(key, box) -> bool:
    if key.number is None:
        return False
    if len(box.number) == 0:
        return True  # no info: try it
    return key.number in box.number
\end{lstlisting}
\vspace{0.3em}

\begin{lstlisting}[language=Python, basicstyle=\ttfamily\scriptsize,
  frame=single, title={\small\itshape Trial  18: True rule}]
def predict(key, box) -> bool:
    # Key number must match the number
    # of shapes on the box.
    return (key.number is not None and
            key.number in box.number)
\end{lstlisting}

\end{minipage}

\caption{Representative hypothesis trajectories generated by LLM-PS-S (left)  and LLM-PS-P (right).}
\label{fig:hypothesis_trajectories}
\end{figure}

\subsection{LLM-PS-P (unreliable evidence, partially observable)}

\textbf{System Prompt}
\begin{verbatim}
You are an intelligent agent playing a game. 
Your task is to open 5 boxes using 13 keys in the fewest attempts. 
You do not need special skills to play this game. 
This game can be played by an 8-12 year old child.
\end{verbatim}

\textbf{User Prompt}
\begin{verbatim}
For each box there is a key that opens it.
The goal of the game is to find the right key for each box, using as few actions as possible.
You have a demonstration video from a teacher telling you how to open all boxes. 
In the video, the teacher says:
"I'm going to show you the right way to unlock the boxes. 
To open the boxes, you have to use a key that matches the color of the box. 
So, to open this red box, I'm going to use this red key. 
Great, now you can open all the doors!”

Each key has an identifier (id) and a color. 
Each key also has either a number or a shape, but not both.

Each box has a color. It also has a shape, which is printed on at least one of its faces.
Not all faces are visible to you initially, but the game allows you to pick up boxes and 
examine them to get more information.

Here are the boxes, lined up in this order:
The first box is red, has a moon shape.
The second box is pink, has a cloud shape.
The third box is white, has a diamond shape.
The fourth box is purple, has a heart shape.
The fifth box is blue, has a triangle shape.

Here are the 13 keys (in no specific order):
The red key is red and has the number 1.
The pink key is pink and has the number 6.
The grey2 key is grey and has the number 2.
The cloud key is grey and has a cloud shape.
The orange4 key is orange and has the number 4.
The green3 key is green  and has the number 3.
The blue key is blue  and has a star shape.
The yellow5 key is yellow and has the number 5.
The heart key is green and has a heart shape.
The white key is white and has the number 7.
The triangle key is yellow and has a triangle shape.
The diamond key is orange and has a diamond shape.
The purple key is purple and has an arrow shape.

You can interact with this environment by taking two types of actions: 
(1) Attempt Action: write Python code that will be used to generate opening attempts, or 
(2) Observe Action: request more information about a given box.

To Take the Observe Action, your output should be exactly 
"PICK UP x", where x is the box id (do not use any other attributes of the box)
    
To take an Attempt Action, you will need to write a Python function that specifies 
a hypothesis about which keys open which boxes.
Your output should use the given starter code, and complete the function called predict 
according to its signature. 
Your output should contain only the Python program for predict, absolutely nothing else. 
It should NOT contain the Key or Box classes.

Here is the history of actions taken and observed evidence. 
Please use them to make your decision:

Open box {BOX1} with key {KEY1}: {OUTCOME1}
Examine {BOX2}: {BOX2.number (real)} faces have shapes on them
Open box {BOX3} with key {KEY3}: {OUTCOME3}
...

Here is the starter code for attempt action

def predict(key, box) -> bool:
    # key is a Key object
    # box is a Box object
    # fill in your code

class Key:
    def __init__(self, id: str, color: str, number: Optional[int], shape: Optional[str]):
        self.id = id
        self.color = color
        self.number = number
        self.shape = shape

class Box:
    def __init__(self, id: str, color: str, shape: str, number: set, position: int):
        self.id = id
        self.color = color
        self.shape = shape
        self.position = position
        self.number: set
        # Current belief over number of shapes on this box.
        # Unknown until PICK UP action is taken.
        # After PICK UP: collapses to a singleton set containing the exact value.
        # In predict(), handle both uncertain and certain cases.
                
Here are the accurate data for all keys and boxes. 
The indexing of the arrays corresponds to the line-up order of boxes.

key_data = {
    "id": ["red", "pink", "grey2", "cloud", "orange4", "green3", "blue", "yellow5", 
            "heart", "white", "triangle", "diamond", "purple"],
    "color": ["red", "pink", "grey", "grey", "orange", "green", "blue", "yellow", 
            "green", "white", "yellow", "orange", "purple"],
    "number": [1, 6, 2, None, 4, 3, None, 5, None, 7, None, None, None],
    "shape": [None, None, None, "cloud", None, None, "star", None, "heart", None, 
            "triangle", "diamond", "arrow"]
}

box_data = {
    "id": ["red", "pink", "white", "purple", "blue"],
    "color": ["red", "pink", "white", "purple", "blue"],
    "shape": ["moon", "cloud", "diamond", "heart", "triangle"], 
    "number": [set(), set(), set(), set(), set()],
    "position": [1, 2, 3, 4, 5],
}


Now is your turn. Respond with either Observe or Attempt action.
\end{verbatim}

\textbf{Example of partially correct Final Hypothesis  
\textsc{gpt-5.2}, LLM-PS-P}

Here the agent fails to observe all boxes, and instead produces a lazy hypothesis that uses the correct rule, but only for boxes that it observed.

\begin{verbatim}
def predict(key, box) -> bool:
    if key.number is not None:
        if hasattr(box, "number") and isinstance(box.number, set) and len(box.number) > 0:
            return key.number in box.number
    # match by color when number evidence isn't available.
    return key.color == box.color
\end{verbatim}

\subsection{LLM-PS-S (unreliable evidence, fully observable)}

\textbf{System Prompt}
\begin{verbatim}
You are an intelligent agent playing a game. 
Your task is to open 5 boxes using 13 keys in the fewest attempts. 
You do not need special skills to play this game. 
This game can be played by an 8-12 year old child.
\end{verbatim}

\textbf{User Prompt: Context}
\begin{verbatim}
For each box there is a key that opens it.
The goal of the game is to find the right key for each box. 
You have a demonstration video from a teacher telling you how to open all boxes. 
In the video, the teacher says:
"I'm going to show you the right way to unlock the boxes. 
To open the boxes, you have to use a key that matches the color of the box. 
So, to open this red box, I'm going to use this red key. 
Great, now you can open all the doors!”
            
The red box has 1 moon shape. 
The pink box has 2 cloud shapes. Each cloud is numbered from 1 to 2.
The white box has 4 diamond shapes. Each diamond is numbered from 1 to 4.
The purple box has 3 heart shapes. Each heart is numbered from 1 to 3.
The blue box has 5 triangle shapes. Each triangle is numbered from 1 to 5.

Here are the 13 keys (in no specific order):
The red key is red and has the number 1.
The pink key is pink and has the number 6.
The grey2 key is grey and has the number 2.
The cloud key is grey and has a cloud shape.
The orange4 key is orange and has the number 4.
The green3 key is green  and has the number 3.
The blue key is blue  and has a star shape.
The yellow5 key is yellow and has the number 5.
The heart key is green and has a heart shape.
The white key is white and has the number 7.
The triangle key is yellow and has a triangle shape.
The diamond key is orange and has a diamond shape.
The purple key is purple and has an arrow shape.
            
Here is the accurate code data for all keys and boxes in the environment. 
The number attribute of boxes denotes the number of faces with a shape on them.

key_data = {
    "id": ["red", "pink", "grey2", "cloud", "orange4", "green3", "blue", 
            "yellow5", "heart", "white", "triangle", "diamond", "purple"],
    "color": ["red", "pink", "grey", "grey", "orange", "green", "blue", 
            "yellow", "green", "white", "yellow", "orange", "purple"],
    "number": [1, 6, 2, None, 4, 3, None, 5, None, 7, None, None, None],
    "shape": [None, None, None, "cloud", None, None, "star", None, "heart", 
            None, "triangle", "diamond", "arrow"]
}

box_data = {
    "id": ["red", "pink", "white", "purple", "blue"],
    "color": ["red", "pink", "white", "purple", "blue"],
    "shape": ["moon", "cloud", "diamond", "heart", "triangle"], 
    "number": [1, 2, 4, 3, 5],
    "position": [1, 2, 3, 4, 5],
}

A hypothesis is a valid Python program that can be executed 
to predict the outcome of a given key and box.
The Python program should have the following signature:

def predict(key, box) -> bool:
    # key is a Key object
    # box is a Box object
    # fill in your code

The Key and Box objects are defined as follows. 
The default value for optional parameters is None.

class Key:
    def __init__(self, id: str, color: str, number: Optional[int], shape: Optional[str]):
        self.id = id
        self.color = color
        self.number = number
        self.shape = shape
class Box:
    def __init__(self, id: str, color: str, number: int, shape: str):
        self.id = id
        self.color = color
        self.number = number
        self.shape = shape                
\end{verbatim}

\textbf{User Prompt: Generate Hypothesis}
\begin{verbatim}
Now, it is your turn to generate a hypothesis.
Your hypothesis should be a Python program that contains exactly the predict function.
Include the signature exactly as provided, no type hints for arguments.

Your output should contain only the Python program, absolutely nothing else.
Your output should NOT contain the Key or Box classes.

\end{verbatim}

\textbf{User Prompt: Refine Hypothesis}
\begin{verbatim}
Now, your task is to improve and refine an existing hypothesis that 
performs poorly on existing evidence.

This is the hypothesis: {HYPOTHESIS},

Here is the evidence from previous attempts. 
Remember that the keys and boxes are physical objects, so for some probability the 
correct key might fail to open the correct box due to a mechanical failure.

Open box {BOX1} with key {KEY1}: {OUTCOME1}
Open box {BOX2} with key {KEY2}: {OUTCOME2}
...

Generate a new hypothesis.
Your hypothesis should be a Python program that contains exactly the predict function, 
including the provided signature.
Your output should contain only the Python program, absolutely nothing else.
Your output should NOT contain the Key or Box classes.
\end{verbatim}

\textbf{Example of Partially Correct Final Hypothesis,  \textsc{gpt-5.2}, LLM-PS-S}

The agent never forgets the misleading rule, but still successfully completes the task once the true rule has been introduced as part of the hypotheses.

\begin{verbatim}
def predict(key, box) -> bool:
    if key.number is not None:
        return key.number == box.number
    return key.color == box.color

\end{verbatim}

\textbf{Examples of Incorrect Final Hypotheses: \textsc{DeepSeek V3.2}, LLM-PS-S}

\begin{verbatim}
def predict(key, box) -> bool:
    if key.shape is not None and key.shape == box.shape:
        return True
    if key.number is not None and key.number == box.number:
        return True
    return False
\end{verbatim}

\begin{verbatim}
def predict(key, box) -> bool:
    if key.shape is not None:
        if key.shape == "cloud" and box.shape == "cloud":
            return True
        if key.shape == "heart" and box.shape == "heart":
            return True
        if key.shape == "triangle" and box.shape == "triangle":
            return True
        if key.shape == "diamond" and box.shape == "diamond":
            return True
        if key.shape == "star" and box.shape == "moon":
            return False
        if key.shape == "arrow" and box.shape == "moon":
            return False
    if key.number is not None:
        if key.number == box.number:
            return True
    return False
\end{verbatim}

\subsection{LLM-PS (reliable evidence, fully observable)}

\textbf{System Prompt}
\begin{verbatim}
You are an intelligent agent playing a game. 
Your task is to open 5 boxes using 13 keys in fewest attempts. 
You do not need special skills to play this game. 
This game can be played by an 8-12 year old child.
\end{verbatim}

\textbf{User Prompt: Context}
\begin{verbatim}
For each box there is a key that opens it.
The goal of the game is to find the right key for each box. 
You have a demonstration video from a teacher telling you how to open all boxes. 
In the video, the teacher says:
"I'm going to show you the right way to unlock the boxes. 
To open the boxes, you have to use a key that matches the color of the box. 
So, to open this red box, I'm going to use this red key. 
Great, now you can open all the doors!”
            
The red box has 1 moon shape. 
The pink box has 2 cloud shapes. Each cloud is numbered from 1 to 2.
The white box has 4 diamond shapes. Each diamond is numbered from 1 to 4.
The purple box has 3 heart shapes. Each heart is numbered from 1 to 3.
The blue box has 5 triangle shapes. Each triangle is numbered from 1 to 5.

Here are the 13 keys (in no specific order):
The red key is red and has the number 1.
The pink key is pink and has the number 6.
The grey2 key is grey and has the number 2.
The cloud key is grey and has a cloud shape.
The orange4 key is orange and has the number 4.
The green3 key is green  and has the number 3.
The blue key is blue  and has a star shape.
The yellow5 key is yellow and has the number 5.
The heart key is green and has a heart shape.
The white key is white and has the number 7.
The triangle key is yellow and has a triangle shape.
The diamond key is orange and has a diamond shape.
The purple key is purple and has an arrow shape.
            
Here is the accurate code data for all keys and boxes in the environment. 
The number attribute of boxes denotes the number of faces with a shape on them.


key_data = {
    "id": ["red", "pink", "grey2", "cloud", "orange4", "green3", "blue", 
            "yellow5", "heart", "white", "triangle", "diamond", "purple"],
    "color": ["red", "pink", "grey", "grey", "orange", "green", "blue", 
            "yellow", "green", "white", "yellow", "orange", "purple"],
    "number": [1, 6, 2, None, 4, 3, None, 5, None, 7, None, None, None],
    "shape": [None, None, None, "cloud", None, None, "star", None, "heart", 
            None, "triangle", "diamond", "arrow"]
}

box_data = {
    "id": ["red", "pink", "white", "purple", "blue"],
    "color": ["red", "pink", "white", "purple", "blue"],
    "shape": ["moon", "cloud", "diamond", "heart", "triangle"], 
    "number": [1, 2, 4, 3, 5],
    "position": [1, 2, 3, 4, 5],
}

A hypothesis is a valid Python program that can be executed 
to predict the outcome of a given key and box.
The Python program should have the following signature:

def predict(key, box) -> bool:
    # key is a Key object
    # box is a Box object
    # fill in your code

The Key and Box objects are defined as follows. 
The default value for optional parameters is None.

class Key:
    def __init__(self, id: str, color: str, number: Optional[int], shape: Optional[str]):
        self.id = id
        self.color = color
        self.number = number
        self.shape = shape
class Box:
    def __init__(self, id: str, color: str, number: int, shape: str):
        self.id = id
        self.color = color
        self.number = number
        self.shape = shape                
\end{verbatim}

\textbf{User Prompt: Generate Hypothesis}
\begin{verbatim}
Now, it is your turn to generate a hypothesis.
Your hypothesis should be a Python program that contains exactly the predict function.
Include the signature exactly as provided, no type hints for arguments.

Your output should contain only the Python program, absolutely nothing else.
Your output should NOT contain the Key or Box classes.

\end{verbatim}

\textbf{User Prompt: Refine Hypothesis}
\begin{verbatim}
Now, your task is to improve and refine an existing hypothesis that 
performs poorly on existing evidence.

This is the hypothesis: {HYPOTHESIS},

Here is the evidence from previous attempts. 

Open box {BOX1} with key {KEY1}: {OUTCOME1}
Open box {BOX2} with key {KEY2}: {OUTCOME2}
...

Generate a new hypothesis.
Your hypothesis should be a Python program that contains exactly the predict function, 
including the provided signature.
Your output should contain only the Python program, absolutely nothing else.
Your output should NOT contain the Key or Box classes.
\end{verbatim}

\textbf{Example of Incorrect Final Hypothesis (both  \textsc{gpt-5.2} and \textsc{DeepSeek V3.2})}

\begin{verbatim}
def predict(key, box) -> bool:
    # Primary rule supported by evidence: numbered keys open boxes with matching counts
    if key.number is not None:
        return key.number == box.number

    # If no number, try matching the depicted shape
    if key.shape is not None:
        return key.shape == box.shape

    # Fallback: match color (as suggested by the demonstration)
    return key.color == box.color
\end{verbatim}

\subsection{ReAct (fully observable)}
\textbf{System Prompt}
\begin{verbatim}
You are an intelligent agent playing a game. 
Your task is to open 5 boxes using 13 keys in fewest attempts. 
You do not need special skills to play this game. 
This game can be played by an 8-12 year old child.
\end{verbatim}

\textbf{User Prompt}
\begin{verbatim}
For each box there is a key that opens it, so the goal of the game is to find 
the right key for each box. 
You have a demonstration video from a teacher telling you how to open all boxes. 
In the video, the teacher says:
"I’m going to show you the right way to unlock the doors. To open the doors, 
you have to use a key that matches the color of the box. So, to open this red box, 
I’m going to use this red key. Great, now you can open all the doors!”

Here are the boxes, lined up in this order:
The red  box has 1 moon shape. 
The pink box has 2 cloud shapes. 
Each cloud is numbered from 1 to 2.
The white box has 4 diamond shapes. 
Each diamond is numbered from 1 to 4.
The purple box has 3 heart shapes. 
Each heart is numbered from 1 to 3.
The blue box has 5 triangle shapes. 
Each triangle is numbered from 1 to 5.

Here are the 13 keys (in no specific order):
The red1  key is red and has the number 1.
The pink6 key is pink and has the number 6.
The grey2 key is grey and has the number 2.
The greycloud key is grey and has a cloud shape.
The orange4 key is orange and has the number 4.
The green3 key is green  and has the number 3.
The bluestar key is blue  and has a star shape.
The yellow5 key is yellow and has the number 5.
The greenheart key is green and has a  heart shape.
The white7 key is white and has the number 7.
The triangleyellow key is yellow and has a triangle shape.
The diamondorange key is orange  and has a diamond shape.
The purplearrow key is purple and has an arrow shape.

You have already tried the following keys-box combinations:
[BOX, KEY, OUTCOME]

Now is your turn to open the boxes. 
Respond in the format "key, box" (e.g. "red1, red") and do not include any other
text in the response.
\end{verbatim}

\clearpage

\subsection*{Example LLM-PS-P Run Log}\label{si:example_log}

Below we show the first 11 trials from a representative run log of a single LLM-PS-P episode (\textsc{gpt-5.2} with low reasoning). The log shows the 
interleaving of \textsc{Attempt} and \textsc{Observe} actions, and the evolution of hypotheses across trials.

\begin{center}
\small
\begin{tabular}{p{0.04\linewidth} p{0.08\linewidth} p{0.82\linewidth}}
\hline
\textbf{$t$} & \textbf{Action} & \textbf{Detail} \\
\hline
1 & Attempt & 
\begin{lstlisting}[language=Python, basicstyle=\ttfamily\scriptsize, frame=none]
def predict(key, box) -> bool:
    # Open a box if the key color matches the box color.
    return key.color == box.color
\end{lstlisting}
\\

 & & \textit{Attempt:} \texttt{white} $\to$ \texttt{white}: \textbf{failure} \\

\hline
2 & Attempt &
\begin{lstlisting}[language=Python, basicstyle=\ttfamily\scriptsize, frame=none]
def predict(key, box) -> bool:
    shape_keys = {'cloud', 'diamond', 'heart', 'triangle'}
    if box.shape in shape_keys:
        return (key.shape is not None) and 
               (key.shape == box.shape)
    # Moon box: fall back to color match
    return key.color == box.color
\end{lstlisting}
\\
 & &  \textit{Attempt:} \texttt{heart} $\to$ \texttt{purple}: \textbf{failure} \\

\hline
3 & Observe & \texttt{PICK UP purple} $\to$ \textit{purple: 3 faces have shape on them.} \\
\hline
4 & Observe & \texttt{PICK UP red} $\to$ \textit{red: 1 face has shape on it.} \\
\hline
5 & Observe & \texttt{PICK UP white} $\to$ \textit{white: 4 faces have shape on them.} \\
\hline
6 & Observe & \texttt{PICK UP pink} $\to$ \textit{pink: 2 faces have shape on them.} \\
\hline
7 & Attempt &
\begin{lstlisting}[language=Python, basicstyle=\ttfamily\scriptsize, frame=none]
def predict(key, box) -> bool:
    # Revert to color matching.
    return key.color == box.color
\end{lstlisting}
\\
 & &  \textit{Attempt:} \texttt{purple} $\to$ \texttt{purple}: \textbf{failure} \\

\hline
8 & Observe & \texttt{PICK UP blue} $\to$ \textit{blue: 5 faces have shape on them.} \\
\hline
9 & Attempt &
\begin{lstlisting}[language=Python, basicstyle=\ttfamily\scriptsize, frame=none]
def predict(key, box) -> bool:
    # Shape counts observed: red=1, pink=2, 
    # purple=3, white=4, blue=5.
    # Key numbers match shape counts exactly.
    shape_counts = {
        'red': 1, 'pink': 2, 'purple': 3,
        'white': 4, 'blue': 5
    }
    return (key.number is not None and
            key.number == shape_counts.get(box.color))
\end{lstlisting}
\\
 & &   \textit{Attempt:} 
\texttt{grey2} $\to$ \texttt{pink}: \textbf{failure} \quad \\
\hline
10 & Attempt &
\begin{lstlisting}[language=Python, basicstyle=\ttfamily\scriptsize, frame=none]
def predict(key, box) -> bool:
    # match key number to number of shapes on the box.
    return (key.number is not None and
            key.number in box.number)
\end{lstlisting}
\\
 & &   \textit{Attempt:} \texttt{red1} $\to$ \texttt{red}: \textbf{success} \quad \\
\hline
11 & Attempt &
\begin{lstlisting}[language=Python, basicstyle=\ttfamily\scriptsize, frame=none]
def predict(key, box) -> bool:
    # match key number to number of shapes on the box.
    return (key.number is not None and
            key.number in box.number)
\end{lstlisting}
\\
 & &   \textit{Attempt:} \texttt{grey2} $\to$ \texttt{pink}: \textbf{success} \quad \\
\hline
\end{tabular}
\end{center}

\vspace{0.5em}
\noindent\textit{Outcome:} Task solved within 17 trials. 5 boxes opened. Correct rule discovered on Trials 9 and 10 

\vspace{0.5em}
\noindent\textbf{Notes.} The agent correctly issues \textsc{Observe} actions to resolve partial observability (trials 3--6, 8), accumulating shape counts. Trial 9 shows a failure due to unreliable evidence when testing a \textit{correct} action.
This episode illustrates both the information-seeking behavior enabled by \textsc{Observe} actions, and LLMs' ability to recover from unreliable actions.

\subsection*{Other LLM failure modes}\label{si:llm_agentfailure}

In pilot experiments we also evaluated several LLM backends not included in the present analysis because of poor program synthesis performance (i.e. gpt-4.1, qwen 3.5). Their failure modes were dominated by:

\begin{itemize}
    \item failure to generate valid executable code 
    \item generating executable code that was not logically valid (i.e. always returns FALSE upon evaluation)
\end{itemize}

Thus, our modeling framework is limited to LLM backends that have the ability to generate executable code consistently based on prompts. 






\section{Model Fitting}\label{si:fitting}

We fit the SoC models to children's task-completion trajectories, represented by the cumulative number of distinct boxes opened at each trial. For the full model, the fitted parameters are
$\theta = (\rho, p_{\mathrm{gen}})$; lesioned variants contain the corresponding subset of these parameters. The prior probability of the true rule, $p_t$, is fixed at $0.02$ as described below.

For each model variant and parameter setting $\theta_i$, we run $R=100$ forward simulations and estimate the probability that the model has opened
$n \in \{0,1,2,3,4,5\}$ distinct boxes by trial $t$:
\[
    \hat{P}(n \mid t,\, \theta_i).
\]

Let $n_{j,t}$ denote the cumulative number of distinct boxes opened by child $j$ by trial $t$. We score parameter setting $\theta_i$ for child $j$ using the simulation-based composite log-likelihood
\[
    \ell(j,\, \theta_i)
    =
    \sum_{t=1}^{T_j}
    \log
    \max\!\left(
        \hat{P}(n_{j,t} \mid t,\, \theta_i),
        \varepsilon
    \right),
\]
where $T_j$ is the number of \textcolor{black}{ATTEMPT} trials recorded for child $j$ and
$\varepsilon = 10^{-10}$ is a probability floor used to avoid taking the logarithm of zero when an outcome is not observed among the $R=100$ simulations. \textcolor{black}{Here, $\log$ denotes the natural logarithm.}

Because cumulative box counts at successive trials are not independent, this quantity is a composite likelihood rather than the exact joint likelihood of the complete action trajectory. It provides a simulation-based measure of how well each parameter setting reproduces each child's task-completion trajectory.

For each model variant with free parameters, we select the setting that maximizes this objective independently for each child:
\[
    \theta_j^*
    =
    \operatorname*{arg\,max}_{\theta_i}
    \ell(j,\, \theta_i).
\]

The parameter grids used in the current fitting are:
\begin{itemize}
    \item \textbf{SoC-Gen}: $p_{gen} \in \{0.1, 0.2, 0.3, 0.4, 0.5, 0.6, 0.7, 0.8, 0.9\}$ (9 settings; $\rho$ lesioned). 
    \item \textbf{SoC-Rel}: $\rho \in \{(1,1), (1,2),  (1,3), (2,1), (3,1), (4,1), (5,1), (6,1), (9,1), (15,1), (19,1)\}$ (11 settings; $p_{gen}$ lesioned). The tuples  represent parameters given for a Beta distribution prior. Note that the subjective reliability parameter $\rho$ is defined in the main text as a scalar in $(0,1]$, corresponding to the expected value of a  $\text{Beta}(\alpha, \beta)$ prior. During fitting, we sweep over $(\alpha, \beta)$ pairs rather than scalar values directly, with  $\mathbb{E}[\rho] = \alpha/(\alpha+\beta)$. The grid thus spans $\mathbb{E}[\rho] \in [0.05, 0.95]$.
    \item \textbf{SoC-Full}: all combinations of the above two grids (99 settings).
    \item \textbf{SoC-Lesioned}: no free parameters, both $p_{gen}, \rho$ lesioned.
\end{itemize}
Parameter rankings were verified to be stable across alternative grid and smoothing choices. The best-fit log-likelihood for child $j$ under a given model variant is $\log \mathcal{L}^*_j = \log \mathcal{L}(j,\, \theta_j^*)$. For SoC-Lesioned, $\log \mathcal{L}^*_j$ is computed directly at the single fixed parameter setting.

The prior probability of the true rule $p_t$ is introduced as a free 
parameter in the main text. In practice, we first swept $p_t \in [0.01, 0.20]$ at 10 levels, and found fitted values to be consistently concentrated near $p_t = 0.02$ across individuals. We therefore fixed $p_t = 0.02$ for all variants in the 
grid search reported here. The results presented here use this fixed value.

\subsection*{Model comparison}

At the individual level, we identify the variant with the highest $\log \mathcal{L}^*_j$ as the best-fitting model. We report the count of children for whom each variant is selected in Figures \ref{fig2}C and \ref{fig:params}. 

At the population level, we compute the mean best-fit log-likelihood across children for each model variant:
\[
    \overline{\log \mathcal{L}}
    \;=\;
    \frac{1}{J}\sum_{j=1}^{J} \log \mathcal{L}^*_j,
\]
with 95\% confidence intervals derived from the standard error of the mean ($\pm 1.96 \times \mathrm{SE}$, where $\mathrm{SE} = \hat{\sigma}/\sqrt{J}$ and $\hat{\sigma}$ is the sample standard deviation across children) (Figure \ref{fig2}B).

Pairwise differences in model fit are tested using paired $t$-tests on $\log \mathcal{L}^*_j$ values (same children scored under each model), with effect sizes reported as Cohen's $d$ for paired samples:
\[
    d \;=\; \frac{\overline{\Delta}}{\hat{\sigma}_\Delta},
\]
where $\overline{\Delta}$ and $\hat{\sigma}_\Delta$ are the mean and standard
deviation of the per-child log-likelihood differences (Table \ref{tab:model_fit}).


\subsection{Supplementary Model Fitting Results}\label{si:model_fit_res}

\textcolor{black}{For robustness, we report model-fitting results for two probability floors $\varepsilon = 10^{-10}$ and $\varepsilon = 0.01$, demonstrating that the model rankings and statistical conclusions remain unchanged.}

\begin{table}[ht]
\centering
\begin{tabular}{lrrrrrrl}
\hline
\textbf{Model} & \textbf{Mean NLL} & \textbf{SD} & \textbf{95\% CI} & \textbf{$t$} & \textbf{$p$} & \textbf{Cohen's $d$} & \\
\hline
SoC-Lesioned & 326.96 & 295.34 & [269.07, 384.84] & 10.144 & $<$.001 & 1.014 & *** \\
SoC-Rel      & 108.61 &  66.25 & [ 95.62, 121.59] & 13.392 & $<$.001 & 1.339 & *** \\
SoC-Gen      & 100.20 & 154.11 & [ 70.00, 130.41] &  3.897 & $<$.001 & 0.390 & *** \\
SoC-Full     &  45.89 &  23.62 & [ 41.26,  50.52] & ---    & ---     & ---   &     \\
\hline
\end{tabular}
\caption{Model comparison across four SoC variants (NLL, lower = better fit). $t$-statistics and Cohen's $d$ are from paired $t$-tests against SoC-Full.  $N = 100$ children. \textcolor{black}{Here probability floor is $\varepsilon = 10^{-10} $.}}
\label{tab:model_fit}
\end{table}

\begin{table}[ht]
\centering
\begin{tabular}{lrrrrrrl}
\hline
\textbf{Model} & \textbf{Mean NLL} & \textbf{SD} & \textbf{95\% CI} & \textbf{$t$} & \textbf{$p$} & \textbf{Cohen's $d$} & \\
\hline
SoC-Lesioned & 101.12 & 63.04 & [ 88.76, 113.47] & 13.046 & $<$.001 & 1.305 & *** \\
SoC-Rel      &  84.99 & 50.62 & [ 75.07,  94.92] & 13.279 & $<$.001 & 1.328 & *** \\
SoC-Gen      &  64.97 & 51.61 & [ 54.86,  75.09] &  6.078 & $<$.001 & 0.608 & *** \\
SoC-Full     &  45.33 & 24.00 & [ 40.63,  50.04] & ---    & ---     & ---   &     \\
\hline
\end{tabular}
\caption{Model comparison across four SoC variants (NLL, lower = better fit). $t$-statistics and Cohen's $d$ are from paired $t$-tests against SoC-Full.  $N = 100$ children. \textcolor{black}{Here probability floor is $\varepsilon = 0.01$} }
\label{tab:model_fit_1}
\end{table}

\begin{figure}[h]
\centering
\begin{subfigure}[b]{0.7\textwidth}
    \includegraphics[width=\textwidth]{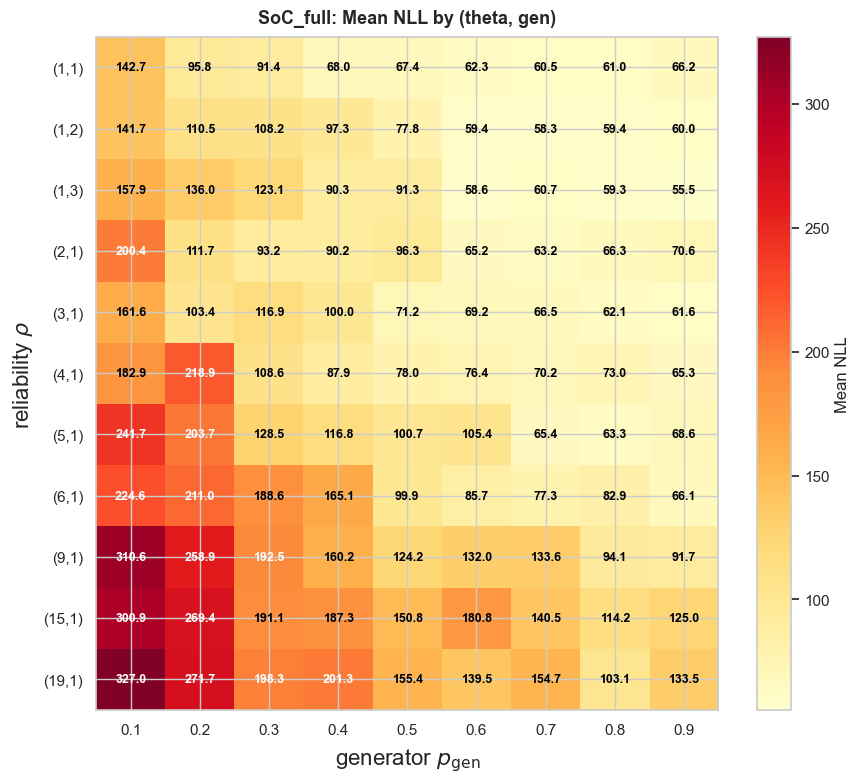}
    \caption{SoC-full}
\end{subfigure}
\hfill
\begin{subfigure}[b]{0.4\textwidth}
    \includegraphics[width=\textwidth]{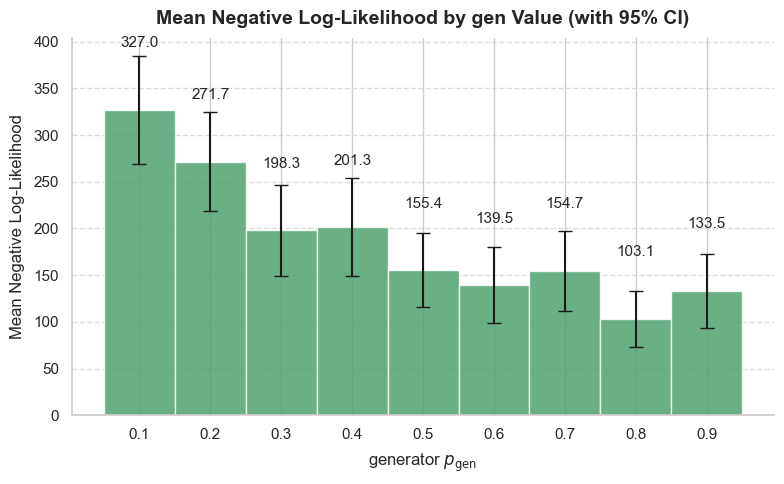}
    \caption{SoC-Gen}
\end{subfigure}
\hfill
\begin{subfigure}[b]{0.4\textwidth}
    \includegraphics[width=\textwidth]{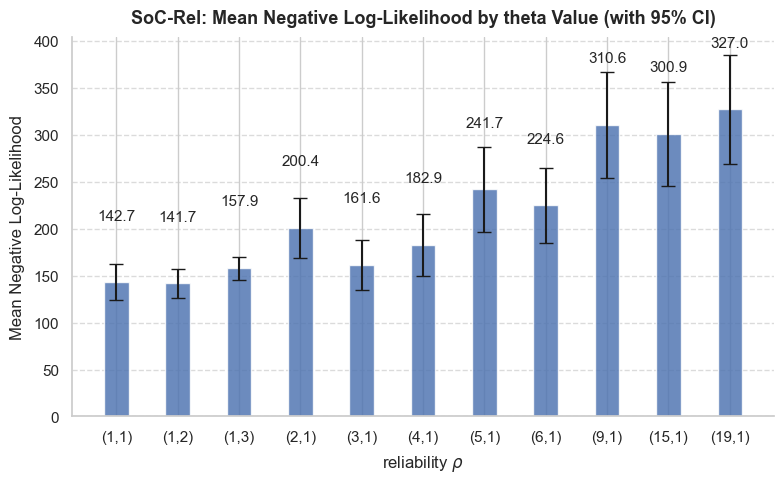}
    \caption{SoC-Rel}
\end{subfigure}
\caption{ Top: SoC-Full parameter fits.  NLL values are represented by color heatmap intensity (brighter = better).  Bottom: Negative Log-Likelihood for each SoC model variant (shorter = better).}
\label{fig:LL_surfaces}
\end{figure}

\begin{figure}[h]
    \centering
    \includegraphics[width=0.6\linewidth]{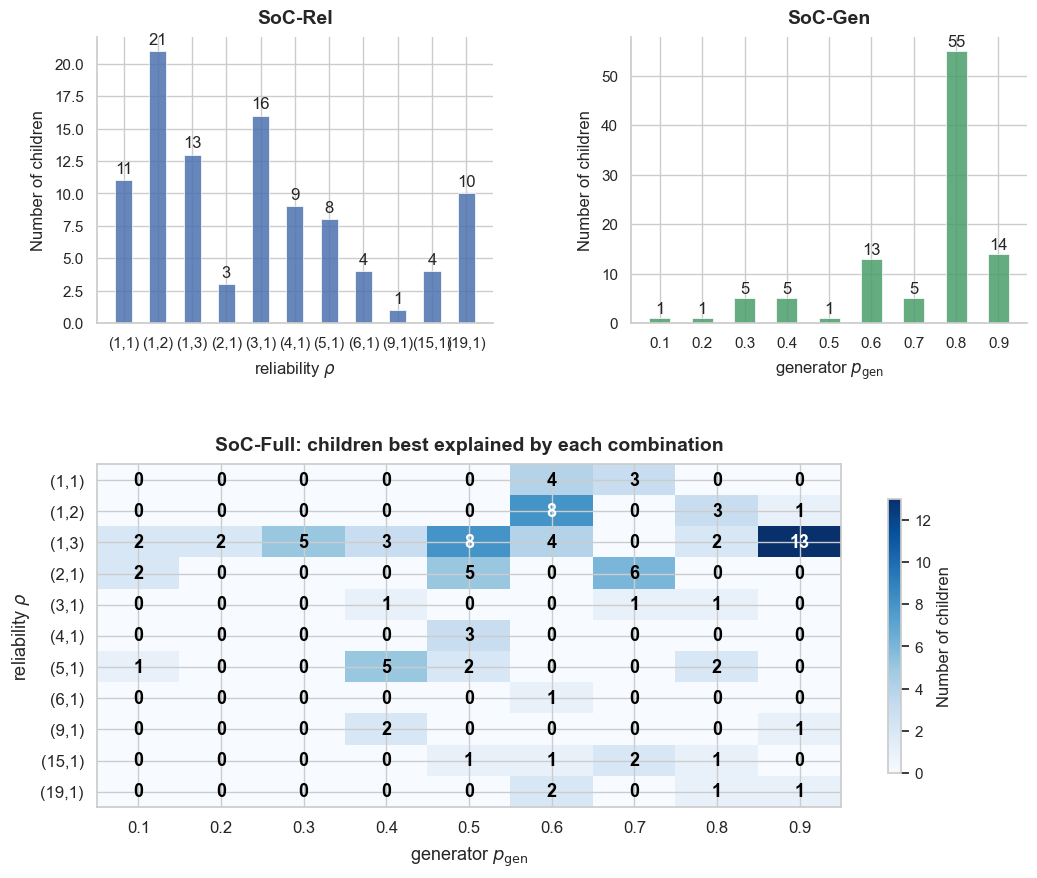}
    \caption{Best-fitting model variant per child.
Top row: Bar plots showing the number of children (out of $N=100$) best explained by each parameter setting, for SoC-Rel and SoC-Gen. Bottom: Heatmap showing the joint distribution of best-fitting parameters for SoC-Full, with two free parameters; cell values indicate the number of children for whom that combination maximised the log-likelihood.}
    \label{fig:params}
\end{figure}

\clearpage

\subsubsection*{Model complexity adjustment}

To account for differences in model complexity, we additionally compare 
models using the Akaike Information Criterion (AIC), computed per child as 
$\text{AIC}_j = - 2\log\mathcal{L}^*_j + 2k$, where $k$ is the number of 
free parameters (SoC-Lesioned: $k=0$; SoC-Rel, SoC-Gen: $k=1$; 
SoC-Full: $k=2$). Mean AIC values across children are reported in 
Table~\ref{tab:aic}.

\begin{table}[ht]
\centering
\begin{tabular}{lrrr}
\hline
\textbf{Model} & \textbf{Mean NLL} & \textbf{$k$} & \textbf{Mean AIC} \\
\hline
SoC-Lesioned & 326.96 & 0 & 653.92 \\
SoC-Rel      & 108.61 & 1 & 219.22 \\
SoC-Gen      & 100.20 & 1 & 202.40 \\
SoC-Full     &  45.89 & 2 &  95.78 \\
\hline
\end{tabular}
\caption{Model comparison by mean AIC (lower is better). The complexity 
penalty for SoC-Full ($2k = 4$) is negligible relative to the NLL 
differences, confirming that the fit improvement of the full model 
outweighs its additional free parameters.}
\label{tab:aic}
\end{table}

\subsection{Relationship between fitted model parameters and age.} \label{si:age}


Fitted model parameters were not significantly associated with child age after
correcting for multiple comparisons (all $p > 0.0125$, Bonferroni-corrected for
four tests), though SoC-full's reliability parameter showed a nominal,
uncorrected correlation ($r = 0.230$, $p = 0.021$; see SI Table~\ref{tab:age-correlations} below).

For each of the 100 children, we took the best-fitting parameter value(s) of each candidate model — the reliability parameter $\rho$ (expressed as the expectation of its fitted Beta prior, $a/(a+b)$) and $p_{\mathrm{gen}}$ for SoC-Rel, SoC-Gen, and SoC-full — and correlated each against child age (in years) using Pearson and Spearman correlations (Table~\ref{tab:age-correlations}). Of the four fitted parameters tested, only the reliability parameter of the full model, SoC-full's $\rho$, showed a statistically reliable relationship with age (Pearson $r = 0.230$, $p = 0.021$; Spearman $\rho = 0.237$, $p = 0.018$), such that older children were better fit by higher assumed key reliability. The remaining parameters (SoC-Rel's $\rho$, SoC-Gen's and SoC-full's $p_{\mathrm{gen}}$) showed weak, non-significant associations with age, with the exception of SoC-Gen's $p_{\mathrm{gen}}$, which reached significance only under the Spearman rank correlation ($\rho = -0.205$, $p = 0.040$) and not the Pearson correlation ($r = -0.159$, $p = 0.114$). We note, however, that the SoC-full reliability effect does not survive a Bonferroni correction for the four comparisons ($\alpha = 0.0125$).

\begin{table}[htbp]
    \centering
    \begin{tabular}{lrrrr}
        \toprule
        Parameter & Pearson $r$ & $p$ & Spearman $\rho$ & $p$ \\
        \midrule
        SoC-Rel $\rho$ (theta expectation)   & 0.177  & 0.078 & 0.133  & 0.186 \\
        SoC-Gen $p_{\mathrm{gen}}$           & -0.159 & 0.114 & -0.205 & 0.040 \\
        SoC-full $\rho$ (theta expectation) & 0.230  & 0.021 & 0.237  & 0.018 \\
        SoC-full $p_{\mathrm{gen}}$         & -0.030 & 0.764 & -0.065 & 0.518 \\
        \bottomrule
    \end{tabular}
    \caption{Correlations between each model's best-fitting parameter(s) and child age ($N = 100$). $\rho$ (reliability) is reported as the expectation of the fitted Beta prior, $a/(a+b)$; $p_{\mathrm{gen}}$ is the generator parameter.}
    \label{tab:age-correlations}
\end{table}

\section{Action-Level Validation of Fitted Model Parameters}~\label{si:action_validation}

\textcolor{black}{We conducted an additional analysis to evaluate whether the SoC models predict children's individual key--box choices, rather than only their cumulative task-completion trajectories. For each child and model variant, we retained the best-fitting parameters obtained from the original model-fitting analysis, without refitting. For each child $j$ and model variant $m$, we performed $R = 100$ 
independent Monte Carlo repetitions of the SMC-S simulation.
At each trial $t$, before supplying the child's observed action,
we recorded the action $\hat{a}_{j,t}^{(r)}$ proposed by the
model under its current particle population. We did not execute
this proposed action. Instead, we supplied the child's actual
key--box choice $a_{j,t}$ and its observed outcome (success or
failure), and updated the particle population using the normal
SMC-S inference procedure. Thus, all repetitions received the
same evidence history, while potentially maintaining different
hypothesis populations and proposing different actions.
We estimated the predictive probability assigned to the child's
observed action as:}

\begin{equation}
\hat{P}_{j,m,t}(a_{j,t}) =
\frac{1}{R}\sum_{r=1}^{R}
\mathbb{I}\left[\hat{a}_{j,t}^{(r)}=a_{j,t}\right].
\end{equation}

\textcolor{black}{For each child and model variant, we calculated the mean negative
log probability per action:}

\begin{equation}
\mathrm{NLL}_{j,m}^{\mathrm{action}} =
-\frac{1}{T_j}\sum_{t=1}^{T_j}
\log\max\left(
\hat{P}_{j,m,t}(a_{j,t}),\,0.001
\right),
\end{equation}

\textcolor{black}{where $T_j$ denotes the number of observed ATTEMPT actions for
child $j$. The probability floor of $0.001$ was applied
consistently across all model variants.}

\begin{figure}[t]
    \centering
    \includegraphics[width=\linewidth]{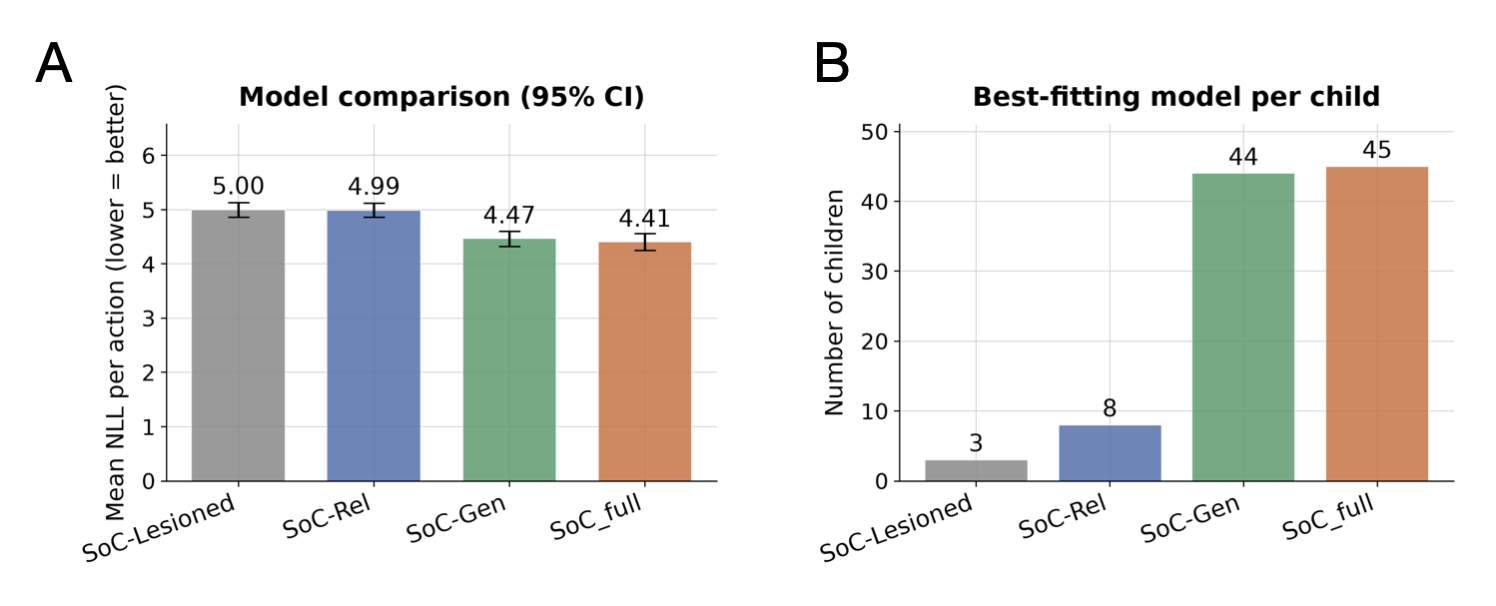}
    \caption{\textbf{Action-level predictive validation of SoC models.}
    \textcolor{black}{(A) Mean negative log probability per observed key--box
    choice for each SoC variant, using parameters previously
    fitted to children's task-completion trajectories.
    Lower values indicate better predictive performance.
    Error bars indicate 95\% confidence intervals across
    children ($N=100$).
    (B) Number of children for whom each model achieved
    the lowest mean negative log probability per action.
    Model predictions were evaluated conditional on each
    child's observed action and outcome history, without
    refitting parameters.}}
    \label{fig:action_validation}
\end{figure}

\textcolor{black}{We report the mean of these per-child scores across the 100
children, with 95\% confidence intervals, and identify the
model assigning the lowest mean negative log score to each child.
SoC-Full achieved the lowest mean NLL per action
(4.405, 95\% CI [4.248, 4.562]), followed by SoC-Gen
(4.466, [4.322, 4.611]), SoC-Rel (4.994, [4.861, 5.126]),
and SoC-Lesioned (5.001, [4.869, 5.134]).
Relative to SoC-Full, mean NLL per action was higher by
0.596 for SoC-Lesioned, 0.589 for SoC-Rel, and 0.061
for SoC-Gen.
SoC-Full provided the best predictions for 45 children,
SoC-Gen for 44, SoC-Rel for 8, and SoC-Lesioned for 3
(Figure~\ref{fig:action_validation}).}

\newpage

\section{Additional Behavioral Results}\label{si:behavior}

Here we show additional analysis of children's behavior in the Box Task, which motivates our model.

\subsection{Observations}

\begin{figure}[th]
    \centering
    \begin{subfigure}[t]{0.58\linewidth}
        \includegraphics[width=\linewidth]{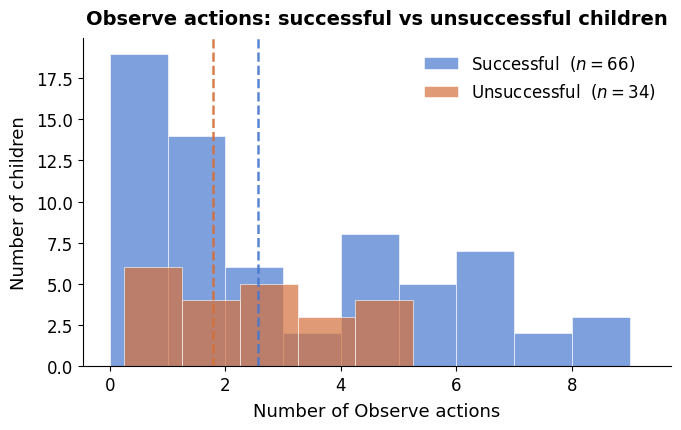}
        \caption{Distribution of \textsc{Observe} actions per child across the full sample ($N = 100$), shown for successful and unsuccessful children.}
    \end{subfigure}
    \hfill
    \begin{subfigure}[t]{0.38\linewidth}
        \includegraphics[width=\linewidth]{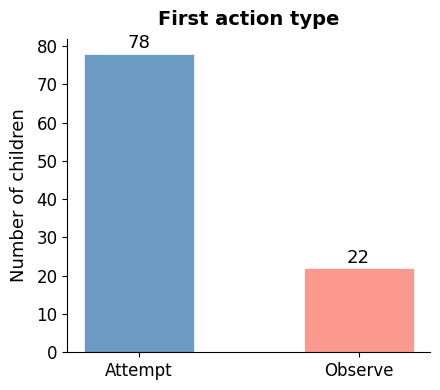}
        \caption{Type of action taken to commence the task}
    \end{subfigure}
    \caption{Behavioral patterns in the Box Task ($N = 100$).}
    \label{fig:si_observe_actions}
\end{figure}

An \textsc{Observe} action is defined as picking up a box to examine its hidden faces. Children vary substantially in how often they seek additional information, with many children making no observations at all (Figure \ref{fig:si_observe_actions}). 
Notably, children who successfully completed the task observed boxes more often (Successful: mean 2.56, SD 2.52;
Unsuccessful: mean 1.79, SD 1.76). Out of 100, 79 children made fewer than 5 Observe actions, implying that not all boxes were observed. This variability motivates the partial observability condition in LLM-PS-P, and contrasts with LLM-PS-P agents, which observe systematically until all boxes are examined.

The number of observations between LLM agents and children is not directly comparable, as observations in LLM-PS-P reveal full information about the box, whcih may not be the case for children. 

\begin{figure}[th]
    \centering
    \includegraphics[width=0.6\linewidth]{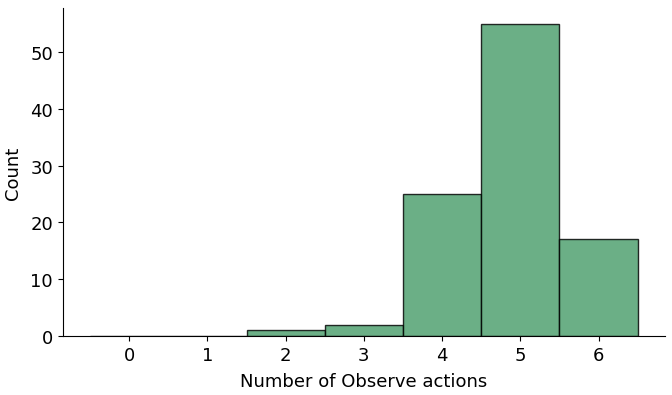}
    \caption{Distribution of \textsc{Observe} actions per simulation ($N = 100$), shown for LLM-PS-P (\textsc{gpt-5.2}, low reasoning). }
    \label{fig:si_observe_actions_agent}
\end{figure}

\subsection{Task Completion}

\begin{figure}[th]
    \centering
    \includegraphics[width=0.6\linewidth]{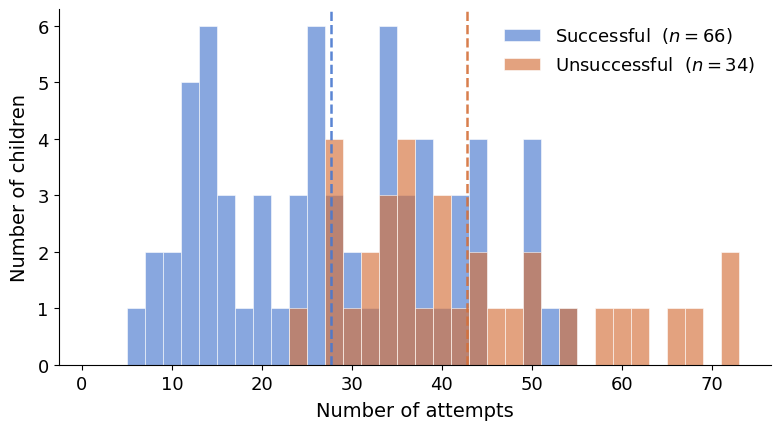}
    \caption{Distribution of attempts at task termination (either all boxes 
    opened or time elapsed) for children who successfully opened all five 
    boxes (blue, $n = 66$) and those who did not (orange, $n = 34$). }
    \label{fig:si_children_completion}
\end{figure}

Children who failed to complete the task accumulated attempts throughout the full 5-minute window without converging on a solution. Such children generally used more attempts compared to successful children. Failure to complete the task consistently reflected reaching a time-out, rather than giving up. 
(Figure \ref{fig:si_children_completion}). 
    
\subsection{Key Reliability}\label{si:rel}

We motivate the \textbf{subjective reliability parameter} $\rho$ in the SoC model by the empirical observation that children consistently re-attempt the same key-box pair on consecutive trials. A repeat is defined as attempting the identical key--box combination  on trial $t+1$ as on trial $t$. 

\begin{figure}[th]
    \centering
    \includegraphics[width=0.7\linewidth]{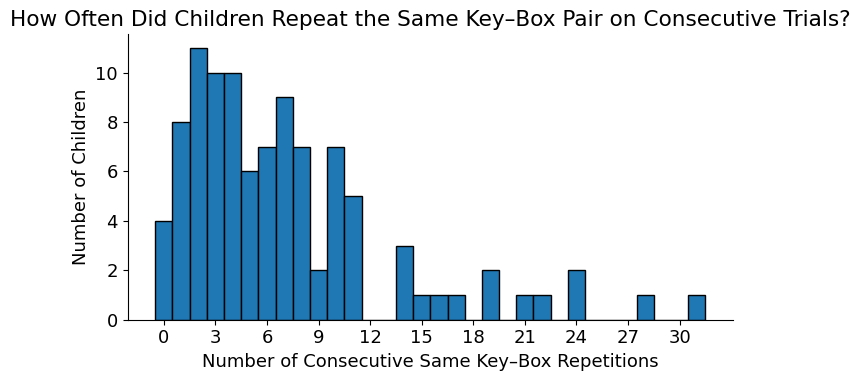}
    \includegraphics[width=0.27\linewidth]{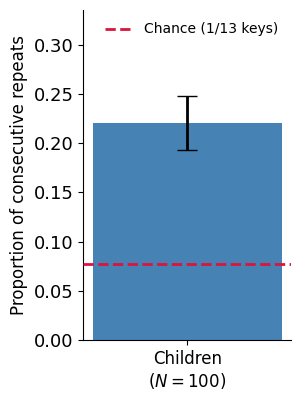}
    \caption{ \textbf{Left}: Number of consecutive trials on which children ($N = 100$) repeated 
    the same key-box pair. \textbf{Right}: Proportion of consecutive trials on which children repeated the same key--box pair, aggregated across participants ($N = 100$). Repetitions above chance are consistent with the subjective reliability account. Error bar indicates 95\% bootstrap confidence intervals 
    across participants.}
    \label{fig:consecutive_repeats}
\end{figure}

We motivate the \textbf{ground-truth reliability parameter} $\rho^*$ used in the LLM-PS-S and LLM-PS-P simulations by the observations that correct key behavior is objectively stochastic in the behavioral data.

\begin{figure}[th]
    \centering
    \includegraphics[width=0.8\linewidth]{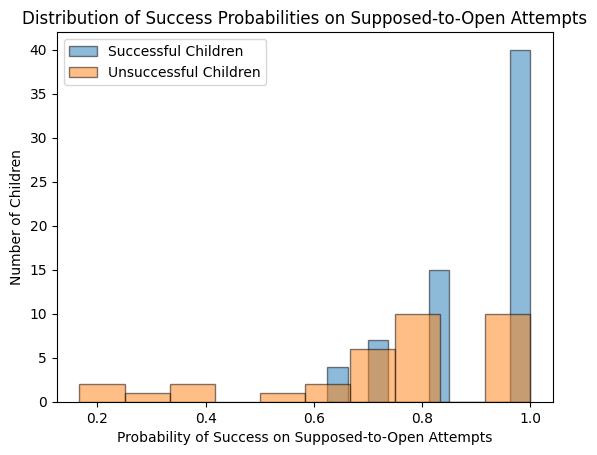}
    \caption{Distribution of ground-truth key reliability in the Box Task, shown separately for children who successfully completed the task (blue) and the ones who did not (orange). }
    \label{fig:key_reliability}
\end{figure}

This is a one-inflated Beta distribution, a mixture of: 

\begin{itemize}
    \item A point mass at exactly 1.0: children who got every single "supposed-to-open" attempt right (50\%, 50 out of 100 children) 
    \item A continuous $Beta(\alpha, \beta)$ component for the rest, describing children with imperfect success rates (50\%, 50 out of 100 children). 
\end{itemize}

MLE fit to the Beta portion estimates \textsc{scipy.stats.beta.fit} : $\alpha=5.9, \beta=2.7$ 

This distribution was used to implement unreliable keys in LLM-PS-S and LLM-PS-P.

\subsection{Generalization} 

\begin{figure}[htbp]
    \centering
    \includegraphics[width=0.7\linewidth]{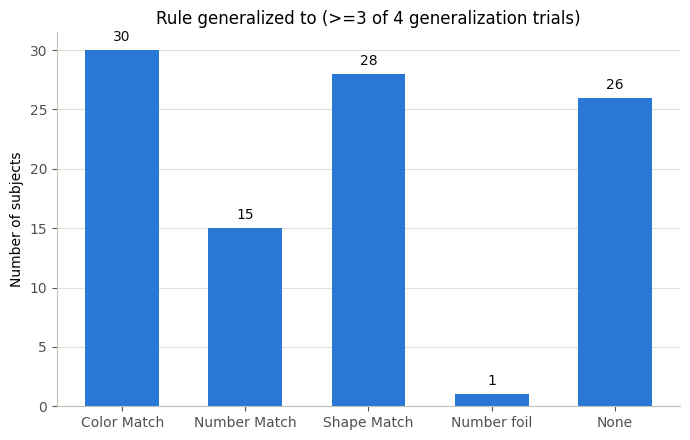}
    \caption{Distribution of the rule each subject generalized to during the four generalization trials. A subject was classified as generalizing to a given rule if choosing the same key type (color match, number match, or shape match) above chance, on at least three of the four trials; subjects who did not show this pattern for any single type were classified as \emph{None}. Of the 100 subjects, 30 generalized to color, 15 to number, 28 to shape, 1 to the number foil, and 26 showed no consistent rule.}
    \label{fig:generalization-histogram}
\end{figure}


\clearpage

\subsection{Initial Hypothesis Selection}

\begin{figure}[th]
    \centering
    \includegraphics[width=0.6\linewidth]{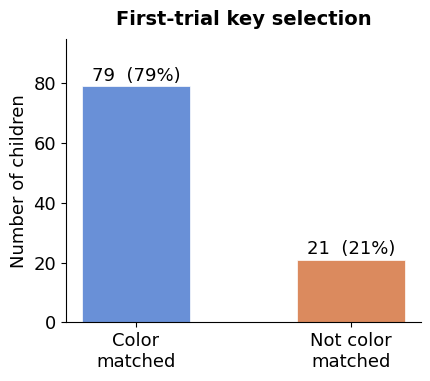}
    \caption{The frequency of first attempt per child across the 
    full sample ($N = 100$) consistent with the taught color-matching rule vs not.}
    \label{fig:si_first_hypothesis}
\end{figure}

\subsection{Attempts subjects made before abandoning a color-matching strategy}

\begin{figure}[htbp]
    \centering
    \includegraphics[width=0.7\linewidth]{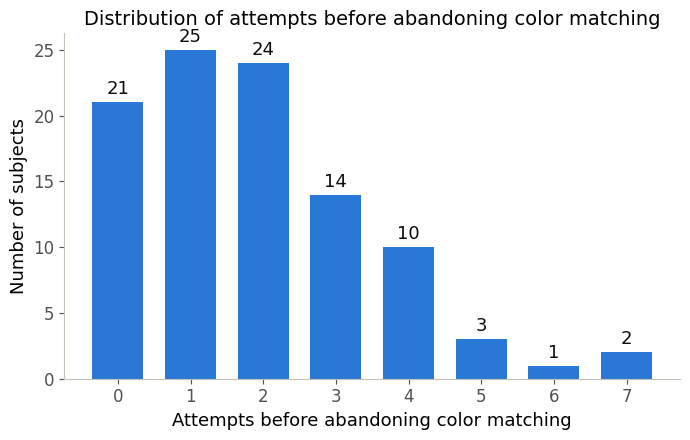}
    \caption{Distribution of the number of attempts subjects made before abandoning a color-matching strategy. For each subject, this counts the leading, consecutive attempts made with a color-matched key, starting from their very first attempt, up to (but not including) their first attempt with a key that did not match the box's color; a value of 0 indicates the subject's first attempt already used a non-color-matched key. Across the 100 subjects, 21 abandoned color matching immediately, while the remainder persisted for 1 to 7 attempts (mode = 1--2 attempts) before switching away from color as a strategy.}
    \label{fig:attempts-before-abandoning-color}
\end{figure}

\subsection{Behavioral Comparison with LLM-Based Agents}
\label{si:llm_behavior}

\begin{figure}[t]
    \centering
    \includegraphics[width=\linewidth]{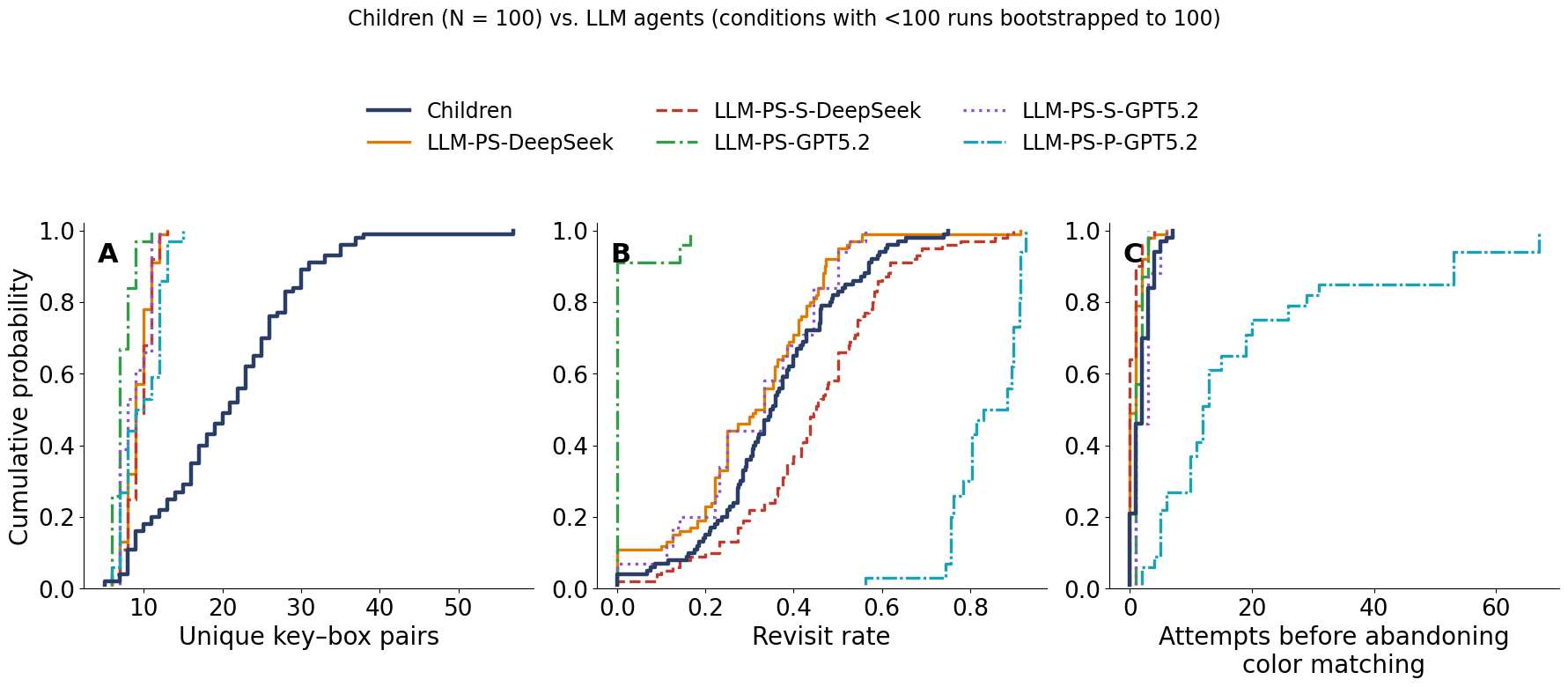}
    \caption{\textbf{Comparison of observed behavior with LLM-based agents.}
    \textcolor{black}{Distributions of (A) unique key--box pairs attempted,
    (B) key--box revisit rate, and (C) attempts before abandoning
    color matching. Children's behavior (dark blue, solid line;
    one value per child, $N=100$) is compared with LLM agents
    (one value per run) in three conditions: reliable keys
    (LLM-PS), stochastic keys (LLM-PS-S), and stochastic keys
    with partial observability (LLM-PS-P).
    LLM-PS and LLM-PS-S are shown for DeepSeek-V3.2
    (100 runs each) and GPT-5.2 (20 runs each);
    LLM-PS-P is shown for GPT-5.2 (20 runs).
    Only key--box attempts are counted; observation actions
    are excluded.}}
    \label{fig:llm_behavior}
\end{figure}

\textcolor{black}{We extend the behavioral analysis in Figure \ref{fig:behavioral_baselines} to LLM-based
agents, comparing the distributions of unique key--box pairs
attempted, key--box revisit rates, and attempts before abandoning color matching (Figure~\ref{fig:llm_behavior}).
These comparisons are descriptive rather than individual-level
fits to children's behavior, as the LLM-PS conditions differ
in evidence reliability and observability. We consider only
ATTEMPT actions, excluding OBSERVE actions from the analysis.}

\clearpage

\section{Code is provided at}
\href{https://anonymous.4open.science/r/Causal-Inference-Under-Uncertainty-BF48/}{https://anonymous.4open.science/r/Causal-Inference-Under-Uncertainty-BF48/}

\section{Computational Resources}\label{si:compute}

SoC simulations were run on standard CPU hardware. Each model variant was simulated 100 times per parameter setting; the full grid search (99 settings for SoC-Full) completed in under 20 hours on a single core. LLM-PS experiments used commercial API access to \textsc{GPT-5.2} (using OpenAI API) and \textsc{DeepSeek V3.2} (using Alibaba cloud API); we have run each variant fewer than 300 times (including prompt debugging and any live experiments); no GPU resources were required for any experiments reported in this paper.

\end{document}